\documentclass[sigconf]{acmart}
\usepackage{subfigure}
\usepackage{bm}
\usepackage{multirow}
\usepackage[switch]{lineno}

\usepackage{microtype}
\usepackage{graphicx}
\usepackage[export]{adjustbox}
\usepackage{subfigure}
\usepackage{booktabs} 
\usepackage{multirow}
\usepackage{amsfonts}
\usepackage{algorithm}
\usepackage{algpseudocode}
\usepackage{amsmath}
\usepackage{array}
\usepackage{dcolumn}
\usepackage{enumitem}
\usepackage{soul}
\usepackage{color}

\newcommand{\x}{{\bm x}}
\newcommand{\xh}{{\bm x}^{\rm h}}
\newcommand{\xp}[1]{{\bm x}^{\rm p}_{#1}}
\newcommand{\be}{{\bm{\epsilon}}}
\newcommand{\xall}{{\bm x}^{\rm all}}
\newcommand{\xnewall}{\widetilde {\bm x} ^ {\rm all}}
\newcommand{\G}{{\mathcal G}}
\newcommand{\bmH}{{\mathbf H}}
\newcommand{\loss}{{\mathcal L}}

\AtBeginDocument{%
	\providecommand\BibTeX{{%
			\normalfont B\kern-0.5em{\scshape i\kern-0.25em b}\kern-0.8em\TeX}}}

\setcopyright{acmcopyright}
\copyrightyear{2023}
\acmYear{2023}
\acmDOI{XXXXXXX.XXXXXXX}

\acmConference[xx]{}{xx}{xx}
\acmPrice{15.00}
\acmISBN{978-1-4503-XXXX-X/18/06}

\begin{document}
	
\title[]{DiffSTG: Probabilistic Spatio-Temporal Graph Forecasting  \\ with Denoising Diffusion Models}

    \author{Haomin Wen$^{1}$, Youfang Lin$^{1}$, Yutong Xia$^{2}$, Huaiyu Wan$^{1}$,  Qingsong Wen$^{3}$, \\ Roger Zimmermann$^{2}$, Yuxuan Liang$^{4, *}$}
    \affiliation{%
		\textsuperscript{\rm 1}School of Computer and Information Technology, Beijing Jiaotong University, China\\
            \textsuperscript{\rm 2}National University of Singapore, \textsuperscript{\rm 3}DAMO Academy, Alibaba Group\\
            \textsuperscript{\rm 4}Hong Kong University of Science and Technology (Guangzhou) *corresponding author\\
            \{wenhaomin, yflin, hywan\}@bjtu.edu.cn;  yutong.xia@u.nus.edu; \\  qingsongedu@gmail.com; rogerz@comp.nus.edu.sg; yuxliang@outlook.com
            \vspace{2em}
         \country{}
	}


 \renewcommand{\shortauthors}{Haomin Wen et al.}

\begin{abstract}
    \par Spatio-temporal graph neural networks (STGNN) have emerged as the dominant model for spatio-temporal graph (STG) forecasting. Despite their success, they fail to model intrinsic \emph{uncertainties} within STG data, which cripples their practicality in downstream tasks for decision-making. To this end, this paper focuses on \emph{probabilistic} STG forecasting, which is challenging due to the difficulty in modeling uncertainties and complex \emph{ST dependencies}. In this study, we present the first attempt to generalize the popular denoising diffusion probabilistic models to STGs, leading to a novel non-autoregressive framework called DiffSTG, along with the first denoising network UGnet for STG in the framework. Our approach combines the spatio-temporal learning capabilities of STGNNs with the uncertainty measurements of diffusion models.   Extensive experiments validate that DiffSTG reduces the Continuous Ranked Probability Score (CRPS) by 4\%-14\%, and Root Mean Squared Error (RMSE) by 2\%-7\% over existing methods on three real-world datasets. The code is in https://github.com/wenhaomin/DiffSTG.
\end{abstract}

	\ccsdesc[500]{Information systems~Data mining}
	\ccsdesc[500]{Applied computing~Forecasting}

	\keywords{Diffusion Model; Probabilistic Forecasting; Spatio-Temporal Graph Forecasting}
	
	\settopmatter{printfolios=true}
	\maketitle

\section{Introduction} \label{introduction}
\par
\begin{figure}[!t]
    \centering
    \includegraphics[width=0.98 \linewidth]{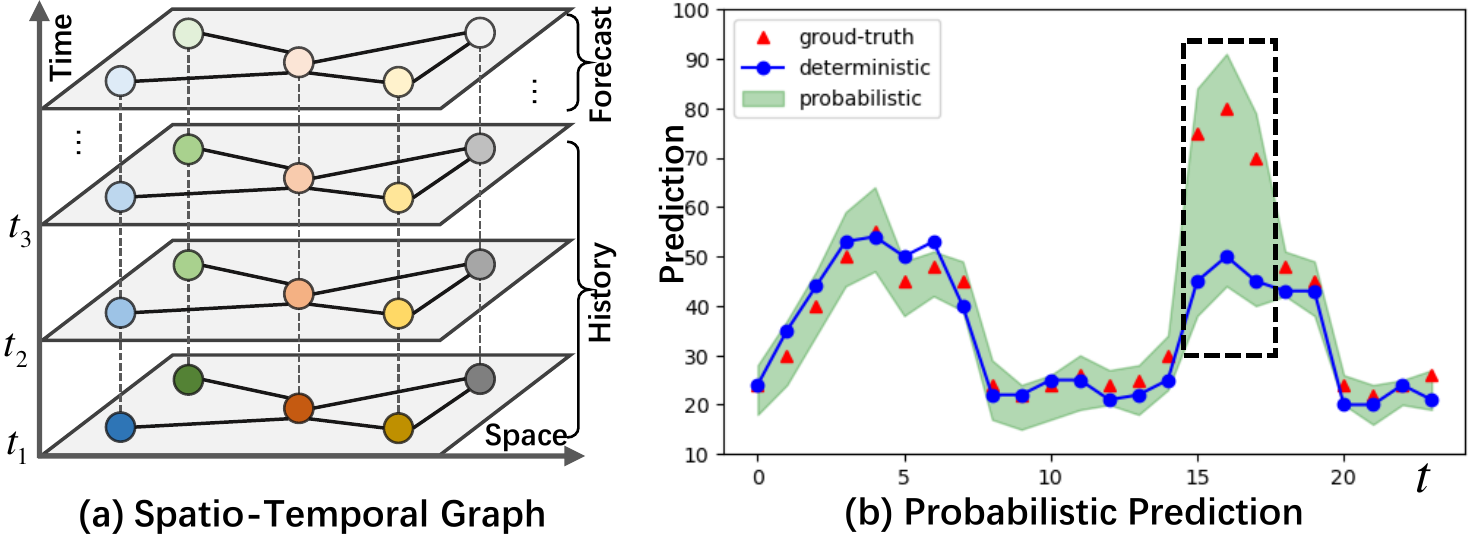}
    \caption{Illustration of probabilistic STG forecasting. (a): Spatio-temporal graph forecasting;  (b): Motivation of probabilistic prediction.}
    \label{fig:probalistic_prediction}
    \vspace{-1em}
\end{figure}

 Humans enter a world that is inherently structured, in which a myriad of elements interact with each other both spatially and temporally, resulting in a spatio-temporal composition. Spatio-Temporal Graph (STG) is the de facto most popular tool for injecting such structural information into the formulation of practical problems, especially in smart cities. In this paper, we focus on the problem of \emph{STG forecasting}, i.e.,  predicting the future signals generated on a graph given its historical observations and the graph structure, such as traffic prediction \cite{li2018diffusion}, weather forecasting \cite{simeunovic2021spatio}, and taxi demand estimation \cite{yao2018deep}. To facilitate understanding, a sample illustration is given in Figure \ref{fig:probalistic_prediction}(a).

\par Recent techniques for STG forecasting are mostly deterministic, calculating future graph signals exactly without the involvement of randomness. Spatio-Temporal Graph Neural Networks (STGNN) have emerged as the dominant model in this research line. They resort to GNNs for modeling spatial correlations among nodes, and Temporal Convolutional Networks (TCN) or Recurrent Neural Networks (RNN) for capturing temporal dependencies. Though promising, these deterministic approaches still fall short of handling \emph{uncertainties} within STGs, which considerably trims down their practicality in downstream tasks for decision-making. For example, Figure \ref{fig:probalistic_prediction}(b) depicts the prediction results of passenger flows in a metro station. In the black box, the deterministic method cannot provide the reliability of its predictions. Conversely, the probabilistic method renders higher uncertainties (see the green shadow), which indicates a potential outbreaking of passenger flows in that region. By knowing a range of possible outcomes we may experience and the likelihood of each, the traffic system is able to take operations in advance for public safety management.

\par While prior endeavors on stochastic STG forecasting were conventionally scarce, we are witnessing a blossom of probabilistic models for time series forecasting \cite{rubanova2019latent,salinas2020deepar,rasul2021autoregressive}. Denoising Diffusion Probabilistic Models (DDPM) \cite{ho2020denoising} are one of the most prevalent methods in this stream, whose key insight is to produce the future samples by \emph{gradually transforming noise into a plausible prediction through a denoising process}. Unlike vanilla unconditional DDPMs that were originally designed for image generation, such transformation function between consecutive steps is conditioned on the historical time series readings. For example, TimeGrad \cite{rasul2021autoregressive} sets the LSTM-encoded representation of the current time series as the condition, and estimates the future regressively. CSDI \cite{tashiro2021csdi} directly utilizes observed values as the condition to model the data distribution.

\par However, the above probabilistic time series models are still insufficient for modeling STGs. Firstly, they only model the temporal dependencies within a single node, without capturing the spatial correlations between different nodes. In reality, objects are correlated with each other spatially, for example, nearby sensors in a road network tend to witness similar traffic trends.  Failing to encode such spatial dependencies will drastically deteriorate the predictive accuracy \cite{STGCN-2018,GraphWaveNet-2019}. Secondly, the training and inference of existing probabilistic time series models, e.g., Latent ODE \cite{rubanova2019latent} and TimeGrad, suffer notorious inefficiency due to their sequential nature, thereby posing a hurdle to long-term forecasting. 

\par To address these issues, we generalize the popular DDPMs to spatio-temporal graphs for the first time, leading to a novel framework called \textbf{DiffSTG}, which couples the spatio-temporal learning capabilities of STGNNs with the uncertainty measurements of DDPMs. Targeting the first challenge, we devise an effective module (UGnet) as the denoising network of DiffSTG. As its name suggests, UGnet leverages a \underline{U}net-based architecture \cite{ronneberger2015u} to capture multi-scale temporal dependencies and \underline{G}NN to model spatial correlations. Compared to existing denoising networks in standard DDPMs, our UGnet performs more accurate denoising in the reverse process by virtue of capturing ST dependencies. To overcome the second issue, our DiffSTG produces future samples in a non-autoregressive fashion. In other words, our framework efficiently generates multi-horizon predictions all at once, rather than producing them step by step as what TimeGrad did. In summary, our contributions lie in three aspects:
\begin{itemize}[leftmargin=*]
    \item We hit the problem of probabilistic STG forecasting from a score-based diffusion perspective with the first shot. Our DiffSTG can effectively model the complex ST dependencies and intrinsic uncertainties within STG data.

    \item We develop a novel denoising network called UGNet dedicated to STGs for the first time. It contributes as a new and powerful member of DDPMs' denoising network family for modeling ST-dependencies in STG data.

    \item We empirically show that DiffSTG reduces the Continuous Ranked Probability Score (CRPS) by 4\%-14\%, and Root Mean Squared Error (RMSE) by 2\%-7\% over existing probabilistic methods on three real-world datasets.

\end{itemize}

\par The rest of this paper is organized as follows. We delineate the concepts of DDPM in Section~\ref{sec:ddpm}. The formulation and implementation of the proposed DiffSTG are detailed in Section~\ref{sec:DiffSTG} and \ref{sec:model_implement}, respectively. We then examine our framework and present the empirical findings in Section~\ref{sec:experiment}. Lastly, we introduce related arts in Section~\ref{sec:related} and conclude in Section~\ref{sec:conclusion}.

\section{Denoising Diffusion Probabilistic Models} \label{sec:ddpm}

\par Given samples from a data distribution $q({\x_0})$, Denoising Diffusion Probabilistic Models (DDPM) \cite{ho2020denoising} are unconditional generative models aiming to learn a model distribution $p_\theta(\x_0)$ that approximates $q({\x_0})$ and is easy to sample from. Let $\x_n$ for $n=1,\cdots, N$ be a sequence of latent variables from the same sample space of $\x_0$ (denoted as $\mathcal X$). DDPM are latent variable models of the form $p_\theta(\x_0)=\int p_\theta(\x_{0:N})d{\x_{1:N}}$. It contains two processes, namely the forward process and the reverse process. 
\par \textbf{Forward Process.} The forward process is defined by a Markov chain which progressively adds Gaussian noise to the observation $\x_0$:
\begin{equation}
    q(\x_{1:N}  | \x_0 ) = {\prod_{n=1}^{N}} q(\x_n | \x_{n-1}),
    \label{eq:forward_process}
\end{equation}
where $q(\x_n | \x_{n-1})$ is a Gaussian distribution as
\begin{equation}
    q(\x_n | \x_{n-1}) = {\mathcal{N}(\x_n;\sqrt{1-\beta_n} \x_{n-1}, \beta_n \mathbf{I} )},
    \label{eq:forward_Gaussian}
\end{equation}
and $\{ \beta_1, \cdots, \beta_N \}$ is an increasing variance schedule  with $\beta_n \in (0,1)$ that represents the noise level at forward step $n$. Unlike typical latent variable models such as the variational autoencoder \cite{rezende2014stochastic}, the approximate posterior $q(\x_{1:N}  | \x_0 )$ in diffusion probabilistic models is not trainable but fixed to a Markow chain depicted by the above Gaussian transition process. 
\par Let ${\hat \alpha}_n = 1 - \beta_n$ and $\alpha_n= \prod_{i=1}^{n} {\hat \alpha}_i$ be the cumulative product of ${\hat \alpha}_n$, a special property of the forward process is that the distribution of $\x_n$ given $\x_0$ has a close form:
\begin{equation}
    q(\x_n|\x_0) = \mathcal{N}(\x_n; \sqrt{\alpha_n}\x_0, (1-\alpha_n){\mathbf{I}}),
    \label{eq:forward_property}
\end{equation}
which can also be expressed as $\x_n = \sqrt{\alpha_n} \x_0 + \sqrt{1 - \alpha_n} {\be}$ by the reparameteriztioin trick \cite{kingma2013auto}, with $\be \in {\mathcal N}(\mathbf{0}; \mathbf{I})$ as a sampled noise. The above property allows us to directly sample $\x_n$ at any arbitrary noise level $n$, instead of computing the forward process step by step.

\par \textbf{Reverse Process.} The reverse process denoises $\x_N$ to recover $\x_0$ recurrently. It also follows a Markov chain but with learnable Gaussian transitions starting with $p(\x_N)={\mathcal{N}(\x_N; \mathbf{0}, \mathbf{I}) }$, which is defined as 
\begin{equation}
    p_{\theta}(\x_{0:N}) = p(\x_N) \prod_{n=N}^{1}p_\theta(\x_{n-1}|\x_n). 
    \label{eq:reverse_process}
\end{equation}
Then, the transition between two nearby latent variables is denoted by 
\begin{equation}
    p_\theta(\x_{n-1}|\x_n)  = \mathcal{N}(\x_{n-1}; {\mu_\theta(\x_n, n), \sigma_\theta(\x_n, n)}), 
    \label{eq:p_sample}
\end{equation}
with shared parameters $\theta$.  Here we choose the same parameterization of $p_\theta(\x_{n-1}|\x_n)$  as in \cite{ho2020denoising} in light of its promising performance on image generation:
\begin{equation}
    {\mu}_\theta (\x_n, n)=\frac{1}{\alpha_n}\left(\x_n-\frac{\beta_n}{\sqrt{1-\alpha_n}} {\bm \epsilon}_\theta\left(\x_n, n\right)\right),
\end{equation}
\begin{equation}
    \sigma_\theta(\x_n, n)= \frac{1 - \alpha_{n-1}}{1 - \alpha_n} \beta_n,
\end{equation}
where ${\be}_\theta({\mathcal X} \times {\mathbb R}) \rightarrow {\mathcal X}$ is a trainable denoising function that decides how much noise should be removed at the current denoising step. The parameters $\theta$ are learned by solving the following optimization problem:
\begin{equation}
\mathop{\min} _\theta \mathcal{L}(\theta)=
\min _\theta \mathbb{E}_{\x_0 \sim q\left(\x_0\right), \be \sim \mathcal{N}(\mathbf{0}, \mathbf{I}), n}\left\|\be-\be_\theta\left(\x_n, n\right)\right\|_2^2.
\label{eq:loss}
\nonumber
\end{equation}
Since we already know $\x_{0}$ in the training stage, and recall that  $\x_n = \sqrt{\alpha_n} \x_0 + \sqrt{1 - \alpha_n} {\be}$ by the property as mentioned in the forward process, the above training objective of unconditional generation can be specified as 
\begin{equation} 
\small
\mathop{\min} _\theta \mathcal{L}(\theta)=
\min _\theta \mathbb{E}\left\|\be-\be_\theta\left({\sqrt{\alpha_n} \x_0 + \sqrt{1 - \alpha_n} {\be}}, n\right)\right\|_2^2.
\label{eq:loss_x0}
\end{equation}
This training objective can be viewed as a simplified version of loss similar to the one in Noise Conditional Score Networks \cite{song2019generative, song2020improved}. Once trained, we can sample $\x_0$ from Eq.~(\ref{eq:reverse_process}) and Eq.~(\ref{eq:p_sample}) starting from the Guassian noise $\x_N$. This reverse process resembles Langevin dynamics, where we first sample from the most noise-perturbed distribution and then reduce the noise scale step by step until we reach the smallest one. We provide details of DDPM in Appendix~\ref{appendix:DDPM_detail}.

\section{DiffSTG Formulation} \label{sec:DiffSTG}

\par Let $\mathcal{G}=\{{\mathcal V}, {\mathcal E}, {\mathbf A} \}$ represent a graph with $V$ nodes, where $\mathcal V$,  $\mathcal E$ are the node set and edge set, respectively. ${\mathbf A} \in {\mathbb R}^{V \times V}$ is a weighted adjacency matrix to describe the graph topology. 
For ${\mathcal V}=\{ v_1, \dots, v_{V} \}$, let ${\mathbf x}_t \in {\mathbb R}^{ F \times V}$ denote $F$-dimentional signals generated by the $V$ nodes at time $t$. Given historical graph signals $\xh = [{\mathbf x_1}, \cdots, {\mathbf x}_{T_h}]$ of $T_h$  time steps and the graph $\G$ as inputs, STG forcasting aims at learning a function $\mathcal F$ to predict future graph signals $\xp{}$, which is formulated as:
\begin{equation}
    {\mathcal F}:(\xh; \G) \rightarrow [{\mathbf x_{T_h + 1}}, \cdots, {\mathbf x_{T_h + T_p}}]:=\xp{},
\end{equation}
where $T_p$ is the forecasting horizon. In this study, we focus on the task of probabilistic STG forecasting, which aims to estimate the distribution of future graph signals.

\par As introduced in Section~\ref{introduction}, on the one hand, current deterministic STGNNs can capture the spatial-temporal correlation in STG data, while failing to model the uncertainty of the prediction. On the other hand, diffusion-based probabilistic time series forecasting models  \cite{rasul2021autoregressive, tashiro2021csdi} have powerful abilities in learning high-dimensional sequential data distributions, while incapable of capturing spatial dependencies and facing efficiency problems when applied to STG data.

\par To this end, we generalize the popular DDPM to spatio-temporal graphs and present a novel framework called DiffSTG for probabilistic STG forecasting in this section. DiffSTG couples the spatio-temporal learning capabilities of STGNNs with the uncertainty measurements of diffusion models.

\subsection{Conditional Diffusion Model}

\par The original DDPM is designed to generate an image from a white noise without condition, which is not aligned with our task where the future signals are generated conditioned on their histories.  Therefore, for STG forecasting, we first extend the DDPM to a conditional one by making a few modifications to the reverse process. In the unconditional DDPM, the reverse process $p_{\theta}(\x_{0:N})$  in Eq.~(\ref{eq:reverse_process}) is used to calculate the final data distribution $q(\x_0)$. To get a conditional diffusion model for our task, a natural approach is to add the history $\xh$ and the graph structure $\G$ as the condition in the reverse process in Eq.~(\ref{eq:reverse_process}). In this way, the conditioned reverse diffusion process can be expressed as
\begin{equation}
    p_{\theta}(\xp{0:N} | \xh, \G) = p(\xp{N}) \prod_{n=N}^{1}p_\theta(\xp{n-1}|\xp{n}, \xh, \G). 
    \label{eq:condition_reverse_process}
\end{equation}
The transition probability of two latent variables in Eq.~(\ref{eq:p_sample}) can be extended as
\begin{equation}
    \begin{array}{l}
        p_\theta(\xp{n-1}|\xp{n}, \xh, \G) \vspace{5pt}  \\ 
        = \mathcal{N}(\xp{n-1}; {\mu_\theta(\xp{n}, n | \xh, \G), \sigma_\theta(\xp{n}, n | \xh, \G)}).
    \end{array}
    \label{eq:condition_p_sample}
\end{equation}
\par Furthermore, the training objective in Eq.~(\ref{eq:loss_x0}) can be rewritten as a conditional one:
\begin{equation}
\mathop{\min} _\theta \mathcal{L}(\theta)=
\min _\theta \mathbb{E}_{\xp{0}, \be} \left\|\be-\be_\theta\left(\xp{n}, n | \xh, \G\right)\right\|_2^2.
\label{eq:condition_loss}
\end{equation}

\subsection{Generalized Conditional Diffusion Model}

\par In Eq.~(\ref{eq:condition_reverse_process})-(\ref{eq:condition_loss}), the condition $\xh$ and denoising target $\xp{}$ are separated into two sample space $\xh \in {\mathcal{X}^{\rm h}}$ and $\xp{} \in {\mathcal{X}^{\rm p}}$. However, they are indeed extracted from two consecutive periods. Here we propose to consider the history $\xh$ and future $\xp{}$ as a whole, i.e., $\xall = [ \xh, \xp{} ] \in \mathbb{R}^{F \times V \times  T}$, where $T = T_h + T_p$. The history can be represented by masking all future time steps in $\xall$, denoted by  $\xall_{\rm msk}$. So that the condition $\xall_{\rm msk}$ and denoise target $\xall$ share the same sample space ${\mathcal{X}^{\rm all}}$. Thus, the masked version of Eq.~(\ref{eq:condition_reverse_process}) can be rewritten as 
\begin{equation}
    \small
    p_{\theta}(\xall_{0:N} | \xall_{\rm msk}, \G) = p(\xall_{N}) \prod_{n=N}^{1}p_\theta(\xall_{n-1}|\xall_{n}, \xall_{\rm msk}, \G). 
    \label{eq:masked_condition_reverse_process}
\end{equation}
The masked version of Eq.~(\ref{eq:condition_loss}) can be rewritten as 
\begin{equation}
\small
\mathop{\min} _\theta \mathcal{L}(\theta)=
\min _\theta \mathbb{E}_{\xall_{0}, \be} \left\|\be-\be_\theta\left(\xall_{n}, n | \xall_{\rm msk}, \G\right)\right\|_2^2.
\label{eq:masked_condition_loss}
\end{equation}
Compared with the formulation in Eq.~(\ref{eq:condition_reverse_process})-(\ref{eq:condition_loss}), this new formulation is a more generalized one which has the following merits. Firstly, the loss in Eq.~(\ref{eq:masked_condition_loss}) unifies the reconstruction of the history and estimation of the future, so that the historical data can be fully utilized to model the data distribution. Secondly, the new formulation unifies various STG tasks in the same framework, including STG prediction, generation, and interpolation \cite{li2004interpolation}.

\par \textbf{Training.} In the training process, we first construct the masked signals $\xall_{\rm msk}$ according to observed values. Then, given the conditional masked information $\xall_{\rm msk}$, graph $\G$ and the target $\xall_{0}$, we sample noise targets $\xall_n = \sqrt{\alpha_n} \xall_0 + \sqrt{1 - \alpha_n} {\be}$, and then train $\epsilon_{\theta}$ by the loss function in Eq.~(\ref{eq:masked_condition_loss}). The training procedure of DiffSTG is presented in Algorithm~\ref{alg:training}.

\floatname{algorithm}{Algorithm}  
\renewcommand{\algorithmicrequire}{\textbf{Input:}}  
\renewcommand{\algorithmicensure}{\textbf{Output:}} 
\begin{algorithm}[htbp]
   \caption{Training of DiffSTG}
   \small
\begin{algorithmic}[1]
   \Require distribution of training data $q(\xall_0)$, number of diffusion step $N$, variance schedule $\{\beta_1, \cdots, \beta_N \}$,  graph  $\G$.
   \Ensure Trained denoising function $\be_{\theta}$ 

   \Repeat
   \State $n \sim {\rm Uniform}(\{ 1, \cdots, N\}), \xall_{0} \sim q(\xall_0)$
   
    \State Constructing the masked signals $\xall_{\rm msk}$ according to observed values
   \State Sample $\be \sim {\mathcal{N}}(\mathbf{0}, \mathbf{I})$ where $\be$'s dimension corresponds to $\xall_{0}$
   \State Calculate noisy targets $\xall_{n}=\sqrt{\alpha_n}\xall_{0} + \sqrt{1 - \alpha_n }\be$

   \State Take gradient step  $\nabla_\theta  \| \be- \be_\theta(\xall_{n}, n | \xall_{\rm msk}, \G) )\|_2^2$ according to Eq.~(\ref{eq:masked_condition_loss})
   \Until{converged}
\end{algorithmic}
\label{alg:training}
\end{algorithm}

\par \textbf{Inference.} As outlined in Algorithm~\ref{alg:samping}, the inference process  utilizes the trained denoising function $\be_{\theta}$ to sample $\xall_{n-1}$ step by step according to Eq.~(\ref{eq:masked_condition_reverse_process}), under the guidance of $\xall_{\rm msk}$ and $\G$. Note that unlike the previous diffusion-based model, i.e., TimeGrad, which requires executing $T_p$ times of reverse diffusion process for predicting the future $T_p$ steps,  DiffSTG only takes one reverse diffusion process to achieve the prediction, thus significantly accelerating the inference speed.

\begin{algorithm}[htbp]
    \caption{Sampling of DiffSTG}
    \small
    \begin{algorithmic}[1]
    \Require Historical graph signal $\xh$,  graph $\G$, trained denoising function $\be_{\theta}$
    \Ensure Future forecasting $\xp{}$

    \State Construct $\xall_{\rm msk}$ according to $\xh$

    \State Sample $\be \sim {\mathcal{N}}(\mathbf{0}, \mathbf{I})$ where $\be$'s dimension corresponds to $\xall_{\rm msk}$
    \For{$n=N$ {\bfseries to} $1$}
        \State Sample $\xall_{n-1}$ using Eq.~(\ref{eq:masked_condition_reverse_process}) by taking $\xall_{\rm msk}$ and $\G$ as condition
    \EndFor
   \State Take out the forecast target in $\xall_{0}$, i.e., $\xp{}$
    \State Return $\xp{}$
    \end{algorithmic}
    \label{alg:samping}
  
\end{algorithm}

\section{DiffSTG Implementation} \label{sec:model_implement}

\begin{figure*}[htbp]
    \centering
    \includegraphics[width=1 \linewidth]{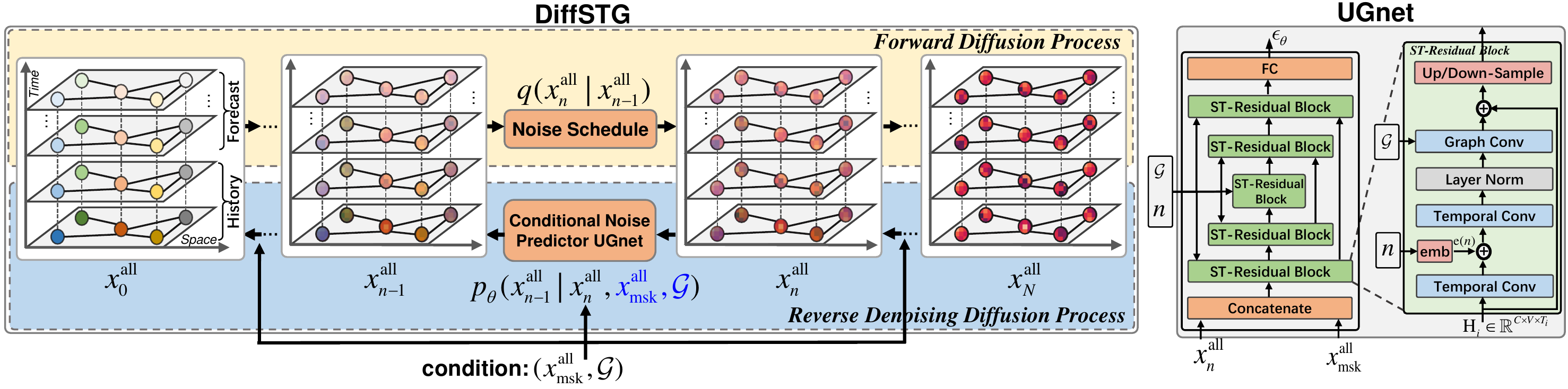} 
    \vspace{-10pt}
    \caption{Illustration of proposed DiffSTG and denoising network UGnet. The inference process of DiffSTG  utilizes the trained denoising function $\be_{\theta}$ (i.e., UGnet) to sample $\xall_{n-1}$ step by step, under the guidance of $\xall_{\rm msk}$ and $\G$. UGnet leverages an Unet-based architecture to capture multi-scale temporal dependencies and the Graph Neural Network (GNN) to model spatial correlations.}
    \vspace{-5pt}
    \label{fig:model}
\end{figure*}

\par After introducing the DiffSTG's formulation, we implement it via elaborately-designed model architecture, which is illustrated in Figure~\ref{fig:model}. At the heart of the model is the proposed denoising network (UGnet) $\be_{\theta}$ in the reverse diffusion process, which performs accurate denoising with the ability to effectively model ST dependencies in the data.

\subsection{Denoising Network: UGnet}
\par  Denoising network $\be_{\theta}$ in previous works can be mainly classified into two classes, Unet-based architecture \cite{ronneberger2015u} for image-related tasks \cite{rombach2022high,voleti2022mcvd, ho2020denoising},
and WaveNet-based architecture \cite{vanwavenet} for sequence-related tasks \cite{kong2020diffwave, liu2022diffsinger, kim2020glow}. These networks consider the input as either grids or segments, lacking the ability to capture spatio-temporal correlations in STG data. To bridge this gap, we propose a new denoising network $\be_\theta({\mathcal X}^{\rm all} \times {\mathbb R} | {\mathcal X}_{\rm msk}^{\rm all}, \G) \rightarrow {\mathcal X}^{\rm all}$, named UGnet. It adopts an \underline{U}net-like architecture in the temporal dimension to capture temporal dependencies at different granularities (e.g., 15 minutes or 30 minutes), and utilizes \underline{G}NN to model the spatial correlations.

\par Specifically, as shown in Figure ~\ref{fig:model}, UGnet takes $\xall_{\rm msk}$, $\xall_{n}$, $n$, $\G$ as inputs, and outputs the denoised noise $\be$. It first concatenates $\xall_{n} \in \mathbb{R}^{F \times V \times T}$ and $\xall_{\rm msk} \in \mathbb{R}^{F \times V \times T}$ in the temporal dimension to form a new tensor $\xnewall_n \in \mathbb{R}^{F \times V \times 2T}$, which is projected to a high-dimensional representation ${\mathbf H} \in {\mathbb R}^{C \times V \times 2T}$ by a linear layer, where $C$ is the projected dimension. Then ${\mathbf H}$ is fed into several Spatio-temporal Residual Blocks (ST-Residual Blocks for short), each capturing temporal and spatial dependencies, respectively. Let ${\mathbf H}_i \in {\mathbb R}^{C \times V \times T_{i}}$ (where ${\mathbf H}_0={\mathbf{H}}$) denote the input of the $i$-th ST-Residual Block, where $T_{i}$ is the length of time dimension.

\par \textbf{Temporal Dependency Modeling.} As shown in Figure ~\ref{fig:epsilon_theta}, at each ST-Residual Block, ${\mathbf H}_i$ is fed into a Temporal Convolution Network (TCN) \cite{bai2018tcn} for modeling temporal dependence, which is a 1-D gated causal convolution of $K$ kernel size with padding to get the same shape with input.  The convolution kernel $\Gamma_{\mathcal T} \in \mathbb{R} ^ {K \times C_{\rm in}^{\rm t} \times C_{\rm out}^{\rm t}}$ maps the input $\mathbf{H}_i$ to outputs ${\mathbf P}_i$, ${\mathbf Q}_i \in \mathbb{R}^{C_{\rm out}^{\rm t} \times V \times T_{i}}$ with the same shape. Formally, the temporal gated convolution can be defined as
\begin{equation}
    \Gamma_{\mathcal T} ({\mathbf H}_{i})  =  {\mathbf P}_i \odot \sigma({\mathbf Q}_i) \in \mathbb{R} ^ {C_{\rm out}^{\rm t} \times V \times T_{i}},
\end{equation}
where $\odot$ is the element-wise Hadamard product, and $\sigma$ is the sigmoid activation function. The item $\sigma({\mathbf Q}_i) $ can be considered a gate that filters the useful information of ${\mathbf P}_i$ into the next layer. We denote the output of TCN as $\overline{\bmH}_i$.

\par \textbf{Spatial Dependency Modeling.} Graph Convolution Networks (GCNs) are generally employed to extract highly meaningful features in the space domain \cite{survey-GNN-2020}. The graph convolution can be formulated as
\begin{equation}
    \Gamma_{\mathcal G}( \overline{\bmH}_i )  =\sigma \left(\Phi \left(\mathbf{A}_{\rm gcn},\overline{\bmH}_i\right) \mathbf{W}_i\right),
\end{equation}
where $\mathbf{W}_i \in \mathbb{R}^{C_{\rm in}^{\rm g} \times C_{\rm in}^{\rm g}}$ denotes a trainable parameter and $\sigma$ is an activation function.
$\Phi(\cdot)$ is an aggregation function that decides the rule of how neighbors' features are aggregated into the target node.  In this work, we do not focus on developing the function $\Phi(\cdot)$. Instead, we use the form in the most popular vanilla GCN~\cite{GCN-kipf2017} that defines a symmetric normalized summation function as $\Phi_{\rm gcn}\left(\mathbf{A}_{\rm gcn},\overline{\bmH}_i\right) = \mathbf{A}_{\rm gcn} \overline{\bmH}_i,$ where $\mathbf{A}_{\rm gcn}=\mathbf{D}^{-\frac{1}{2}}(\mathbf{A}+\mathbf{I})\mathbf{D}^{-\frac{1}{2}} \in \mathbb{R}^{V \times V}$ is a normalized adjacent matrix of graph $\G$. $\mathbf{I}$ is the identity matrix and $\mathbf{D}$ is the diagonal degree matrix with $\mathbf{D}_{ii}=\sum_{j}(\mathbf{A}+\mathbf{I})_{ij}$. Note that we reshape the output of the TCN layer to $\overline{\bmH}_i \in \mathbb{R}^{V \times  C_{\rm in}^{\rm g}}$, where $C_{\rm in}^{\rm g} = T_{i} \times C_{\rm out}^{\rm t}$, and fed this node feature $\overline{\bmH}_i$  to GCN.

\par \textbf{Noise Level Embedding.} As shown in the right part of Figure~\ref{fig:model}, like previous diffusion-based models \cite{rasul2021autoregressive}, we use positional encodings of the noise level $n \in [1,N]$ and process it using a transformer positional embedding \cite{vaswani2017attention}:
\begin{equation}
\small
    \mathbf{e}(n)=\left[\ldots, \cos \left(n / r^{\frac{-2 d}{D}}\right), \sin \left(n / r^{\frac{-2 d}{D}}\right), \ldots\right]^{\mathrm{T}},
\end{equation}
where $d=1,\cdots, D/2$ is the dimension number of the embedding (set to 32), and $r$ is a large constant 10000. For more details about UGnet, please refer to Appendix~\ref{appendix:UGnet}.

\par We highly the unique characteristics and our special design of UGnet by making comparisons with the popular Unet used in diffusion models for the computer version.  1) U-Net is designed for extracting semantic features from Euclidean data, such as images, while UGNet is designed for non-Euclidean STG data to model the spatial-temporal correlations. 2) The U-structure in U-Net serves to aggregate features in the spatial dimension to extract high-level semantic features in the image, while the U-structure in UGNet aggregates signals in the time dimension to capture time dependencies at different granularities. 3) U-Net utilizes 2D-CNNs to capture the spatial correlation, whereas UGNet utilizes graph convolution to model the spatial correlation, which is more suitable for graph-based data.  In summary, the proposed UGnet goes beyond a simple replacement of  U-Net by changing the input from image to graph $\G$,  and involves careful design considerations for STG data.

\subsection{Sampling Acceleration}
\par From a variational perspective, a large $N$ (e.g.,  $N=1000$ in \cite{ho2020denoising}) allows results of the forward process to be close to a Gaussian distribution so that the reverse denoise process started with Gaussian distribution becomes a good approximation. However, large $N$ makes the sampling low-efficiency since all $N$ iterations have to be performed sequentially. To accelerate the sampling process, we adopt the sampling strategy in \cite{song2020denoising}, which only samples a subset $\{\tau_1, \cdots, \tau_{M} \}$ of $M$ diffusion steps. Formally, the accelerated sampling process can be denoted as
\begin{equation}
\begin{array}{l}
    \small
     \x_{\tau_{m-1}}=\sqrt{\alpha_{\tau_{m-1}}} {\left(\frac{\x_{\tau_m}-\sqrt{1-\alpha_{\tau_m}} \be_\theta^{(\tau_m)}}{\sqrt{\alpha_{\tau_m}}}\right)}+ \vspace{5pt} \\ 
     {\sqrt{1-\alpha_{\tau_{m-1}}-\sigma_{\tau_m}^2} \cdot \be_\theta^{(\tau_m)}} + {\sigma_{\tau_m} \be_{\tau_m}}, 
\end{array}
\end{equation}
where $\be_{\tau_m} \sim \mathcal{N}(\mathbf{0}, \mathbf{I})$ is standard Gaussian noise independent of $\x_n$. And $\sigma_{\tau_m}$ controls how stochastic the denoising process is. We set $\sigma_n=\sqrt{\left(1-\alpha_{n-1}\right) /\left(1-\alpha_n\right)} \sqrt{1-\alpha_n / \alpha_{n-1}}$ for all diffusion steps, to make the generative process become a DDPM. When the length of the sampling trajectory is much smaller than $N$, we can achieve significant increases in computational efficiency. Moreover, note that the data in the last $k$ few reverse steps $\xall_{i}$ ($i \in \{1,\dots, k\}$) can be considered a good approximation of the target. Thus we can also add them as samples, reducing the number of the reverse diffusion process from $S$ to $S/k$, where $S$ is the required sample number to form the data distribution.

\subsection{Comparsion among Different Approaches}

\par
\begin{figure}[t]
    \centering
    \includegraphics[width=1 \linewidth]{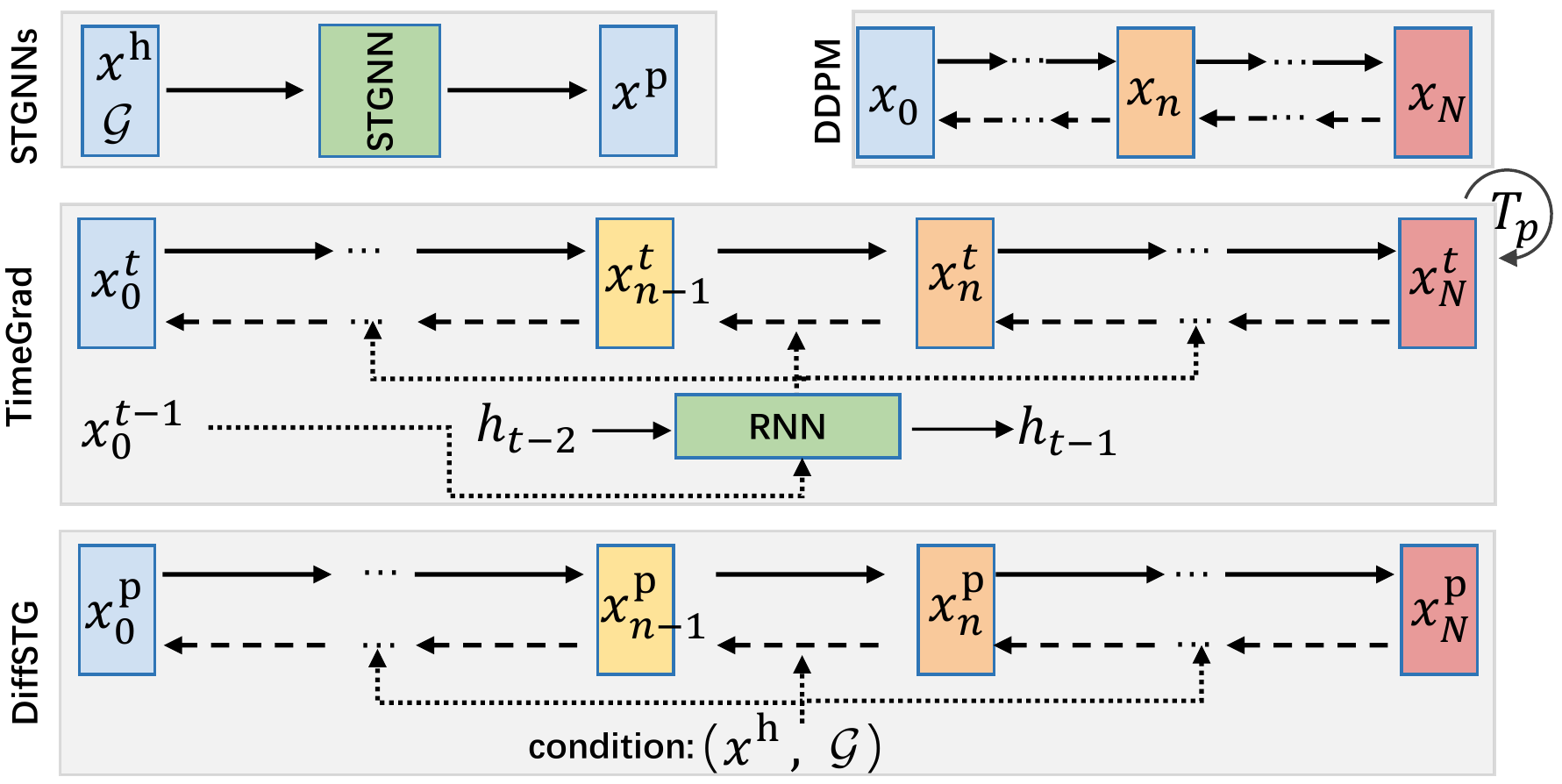} 
    \vspace{-1.5em}
    \caption{Overview of different models.}
    \label{fig:overview}
    \vspace{-15pt}
\end{figure}

 We give the overview of related models in Figure~\ref{fig:overview}: i) Deterministic STGNNs calculate future graph signals exactly without the involvement of randomness. While the vanilla DDPM is a latent variable generative model without condition; ii) To estimate the data distribution from a trained model, i.e., getting $S$ samples, TimeGrad runs $S \times T_p \times N $ diffusion steps for the prediction of all future time steps, where  $N$, $T_p$ is the diffusion step, prediction length, respectively; iii) Compared with current diffusion-based models for time series, DiffSTG 1) incorporates the graph as the condition so that the spatial correlations can be captured, and 2) is a non-autoregressive approach with $\widetilde{S}~\times \widetilde{N} $ diffusion steps to get the estimated data distribution, where $\widetilde{S}=S/k <S$ and $\widetilde{N}=M < N$.

\section{Experiments} \label{sec:experiment}

\par We conduct extensive experiments to evaluate the effectiveness of our proposed DiffSTG on three real-world datasets and compare it with other probabilistic baselines. 

\subsection{Experimental Setup}
\textbf{Datasets.} In the experiments, we choose three real-world datasets from two domains, including a traffic flow dataset PEMS08 \cite{STSGCN-2020}, and two air quality datasets AIR-BJ and AIR-GZ \cite{yi2018deep}.  PEMS08 is a traffic flow dataset collected by the Caltrans Performance Measurement System (PeMS). It records the traffic flow recorded by sensors (nodes) deployed on the road network. In this work, we use the dataset extracted by STSGCN \cite{STSGCN-2020}. The traffic networks (adjacency matrix) for these datasets are constructed according to the actual road network. If the two sensors are on the same road, the two points are considered connected in the spatial network.

\begin{table}[htbp]
    \caption{Details of all datasets.}
    \centering
    \vspace{-1em}
    \setlength \tabcolsep{3pt}
    \resizebox{1 \linewidth}{!}{
	\begin{tabular}{cccccc}
		\toprule
		Dataset & Nodes & $F$  &   Data Type    &     Time interval     &      \#Samples     \\ 
		\midrule
		PEMS08  &  170  &   1    & Traffic flow   &       5 minutes       & 17,856 \\
            AIR-BJ  &  34  &   1    & PM$_{2.5}$   &       1 hour       &  8,760 \\
            AIR-GZ  &  41  &   1    & PM$_{2.5}$   &       1 hour       &  8,760 \\
		\bottomrule
	\end{tabular}
        }
	\label{tab:dataset}
\end{table}

\par The air quality datasets AIR-BJ and AIR-GZ consist of one-year PM$_{2.5}$ readings collected by air quality monitoring stations in two metropolises (i.e., Beijing and Guangzhou) in China \cite{yi2018deep}, respectively. AIR-BJ records data from 34 stations in Beijing from 2019/01/01 to 2019/12/31. And AIR-GZ records data from 41 stations in Guangzhou from 2017/01/01 to 2017/12/31. We build the spatial correlation matrix $\mathbf A$ using the distance between two stations. Statistics of the datasets are shown in Table~\ref{tab:dataset}. 

\begin{table*}[!t]
\centering
\caption{Experiment results of probabilistic methods. Smaller MAE, RMSE, and CRPS indicate better performance. The best result is in bold, and the second-best result is underlined.}
\setlength\tabcolsep{12.2 pt}
\begin{tabular}{l|ccc|ccc|ccc}
\hline 
\multirow{2}{*}{Method}
& \multicolumn{3}{c|}{ AIR-BJ }& \multicolumn{3}{c|}{  AIR-GZ }& \multicolumn{3}{c}{ PEMS08 }\\
\cline{2-10}
 & MAE & RMSE & CRPS & MAE & RMSE & CRPS & MAE & RMSE & CRPS\\
\hline \hline
Latent ODE \cite{rubanova2019latent} & 20.61 & 32.27 & 0.47 & 12.92 & 18.76 & 0.30 & 26.05 & 39.50 & 0.11 \\ \hline
DeepAR \cite{salinas2020deepar} & 20.15 & 32.09 & 0.37 & 11.77 & 17.45 & \underline{0.23} & 21.56 & 33.37 & \underline{0.07} \\ \hline
CSDI \cite{tashiro2021csdi} & 26.52 & 40.33 & 0.50 & 13.75 & 19.40 & 0.28 & 32.11 & 47.40 & 0.11 \\ \hline
TimeGrad \cite{rasul2021autoregressive} & \underline{18.64} & \underline{31.86} & \underline{0.36} & 12.36 & 18.15 & 0.25 & 24.46 & 38.06 & 0.09 \\ \hline
MC Dropout \cite{wu2021quantifying} & 20.80 & 40.54 & 0.45 & \underline{11.12} & \underline{17.07} & 0.25 & \underline{19.01} & \underline{29.35} & 0.07 \\ \hline
DiffSTG (ours) & \textbf{17.88} & \textbf{29.60} & \textbf{0.34} & \textbf{10.95} & \textbf{16.66} & \textbf{0.22} & \textbf{17.68} & \textbf{27.13} & \textbf{0.06} \\ 
Error reduction & -4.1\% & -7.1\% & -5.6\% & -1.5\% & -2.4\% & -4.3\% & -7.0\% & -7.6\% & -14.3\% \\ \hline
\end{tabular}

\label{tab:results}
\end{table*}

\par \noindent \textbf{Baselines.} Both start-of-the-art probabilistic and deterministic models are included for the performance comparison.
\par \textit{Probabilistic baselines.} The following methods are implemented as baselines for probabilistic STG forecasting:

\begin{itemize}[leftmargin=*]
    \item Latent ODE \cite{rubanova2019latent}. It defines a probabilistic generative process over time series from a latent initial state, which can be trained with variational inference.
    \item DeepAR \cite{salinas2020deepar}, which utilizes a Gaussian distribution to model the data distribution; 
    \item TimeGrad \cite{rasul2021autoregressive}, which is an auto-regressive model that combines the diffusion model with an RNN-based encoder; 
    \item CSDI \cite{tashiro2021csdi}, which is a diffusion-based non-autoregressive model first proposed for multivariate time series imputation. We mask all the future signals to adapt CSDI to our task.
    \item MC Dropout \cite{wu2021quantifying}, which is developed based on MC Dropout \cite{gal2017concrete} for probabilistic spatio-temporal forecasting.
\end{itemize}

\par \textit{Deterministic baselines.} We choose some popular and state-of-the-art methods for comparison:
\begin{itemize}[leftmargin=*]

    \item DCRNN \cite{li2018diffusion}: Diffusion Convolutional Recurrent Neural Network integrates diffusion convolution with sequence-to-sequence architecture to learn the representations of spatial dependencies and temporal relations.
    
    \item STGCN \cite{STGCN-2018}: Spatial-Temporal Graph Convolution Network combines spectral graph convolution with 1D convolution to capture spatial and temporal correlations.

    \item STGNCDE \cite{choi2022graph}. Spatio-Temporal Graph Neural Controlled Differential Equation introduces two neural control differential equations (NCDE) for processing
spatial and sequential data, respectively, which can be considered as an NCDE-based interpretation
of graph convolutional networks.

    \item GMSDR \cite{liu2022msdr}: Graph-based Multi-Step Dependency Relation improves RNN by explicitly taking the hidden states of multiple historical time steps as the input of each time unit.

\end{itemize}

\par \noindent \textbf{Metrics.} We choose the Continuous Ranked Probability Score (CRPS) \cite{matheson1976scoring} as the metric to evaluate the performance of probabilistic prediction, which is commonly used to measure the compatibility of an estimated probability distribution $F$ with an observation $x$:
\begin{equation}
    {\rm CRPS}(F, x)=\int_{\mathbb{R}}(F(z)-\mathbb{I}\{x \leq z\})^2 \mathrm{~d} z,
\end{equation}
where $\mathbb{I}\{x \leq z\}$ is an indicator function which equals one if $x \leq z$, and zero otherwise. Smaller CRPS means better performance.

In addition, we leverage Mean Absolute Error (MAE) and Root Mean Squared Error (RMSE) to evaluate the performance of deterministic prediction. Let $Y$ be the label, and $\hat{Y}$ denote the predictive result. 
\begin{equation}
    {\rm MAE}(Y, \hat{Y})=\frac{1}{\left| Y \right|}\sum^{\left| Y \right|}_{i=1}\left| Y_{i} - \hat{Y}_{i}\right|,
\end{equation}
\begin{equation}
   {\rm RMSE}(Y, \hat{Y})=\sqrt{\frac{1}{\left| Y \right|}\sum^{\left| Y \right|}_{i=1}\left( Y_{i} - \hat{Y}_{i}\right)^{2}}.
\end{equation}
where a smaller metric means better performance. Among those metrics, CRPS is the primary metric of our focus to evaluate the performance of different probabilistic methods.

\par \noindent \textbf{Implementation Details.} As for hyperparameters, we set the batch size as 8 and use the Adam optimizer with a learning rate of 0.002, which is halved every 5 epochs. For CSDI and DiffSTG, we adopt the following quadratic schedule for variance schedule: $ \beta_n=\left(\frac{N-n}{N-1} \sqrt{\beta_1}+\frac{n-1}{N-1} \sqrt{\beta_N}\right)^2$. We set the minimum noise level $\beta_1 =  0.0001$ and hidden size $C= 32$ and search the number of the diffusion step $N$ and the maximum noise level $\beta_N$ from a given parameter space ($N \in [50, 100, 200]$, and $\beta_N \in [0,1, 0.2, 0.3, 0.4]$), and each model's best performance is reported in the experiment. For TimeGrad, as TimeGrad is an extremely time-consuming method, we followed the recommendations from the original paper and performed a search in a smaller diffusion setting space for the number of diffusion steps ($N \in [50, 100]$) and the maximum noise level ($\beta_N \in [0,1, 0.2]$). This approach still allowed us to optimize TimeGrad's performance while considering the computational constraints. For other baselines, we utilize their codes and parameters in the original paper. For all datasets, the history length $T_h$, and prediction length $T_p$ are both set to 12. All datasets are split into the training, validation, and test sets in chronological order with a ratio of 6:2:2. The models are trained on the training set and validated on the validation set by the early stopping strategy. 

\subsection{Performance Comparison}

\par The efforts of stochastic models for probabilistic STG forecasting were traditionally scarce. Hence, we compare our model with state-of-the-art baselines in the field of probabilistic time series forecasting, including Latent ODE \cite{rubanova2019latent}, DeepAR \cite{salinas2020deepar}, TimeGrad \cite{rasul2021autoregressive}, CSDI \cite{tashiro2021csdi}, and a recent STG probabilistic forecasting method MC Dropout \cite{wu2021quantifying}. We choose the Continuous Ranked Probability Score (CRPS) \cite{matheson1976scoring} as an evaluation metric. We also report MAE and RMSE of the deterministic forecasting results by averaging $S$ (set to $8$ in our paper) generated samples. Note that CRPS is the primary metric to evaluate the performance of those probabilistic methods. 

\par In Table \ref{tab:results}, DiffSTG outperforms all the probabilistic baselines: it reduces the CRPS by 5.6\%, 4.3\%, and 14.3\% on the three datasets compared to the most competitive baseline in each dataset, respectively. Distributions in DeepAR and Latent ODE can be viewed as some types of low-rank approximations of the target, which naturally restricts their capability to model the true data distribution. TimeGrad outperforms LatentODE due to its DDPM-based architecture with tractable likelihoods that models the distribution in a general fashion. CSDI is a diffusion-based model originally proposed for time series imputation, thus performing worse in our forecasting tasks. MC Dropout achieves the second best performance on MAE and RMSE in most datasets, due to its strong ability in modeling the ST correlations. Our DiffSTG yields the best performance in both deterministic and probabilistic prediction, revealing that it can preserve the spatio-temporal learning capabilities of STGNNs as well as the uncertainty measurements of the diffusion models.

\textbf{Inference Time.} Generally, diffusion-based forecasting models are much slower than other methods (due to their recurrent denoising nature), but they deliver promising performance, as introduced in TimeGrad and DDPM. To this end, here we mainly compare the inference speed of the diffusion-based forecasting method, including DiffSTG, TimeGrad, and CSDI. Table~\ref{tab:time_cost} reports the average time cost per prediction of the three diffusion-based forecasting models. We observe that TimeGrad is extremely time-consuming due to its recurrent architecture. DiffSTG (with $M$=100 and $k$=1) achieves 40$\times$ speed-up compared to TimeGrad, which stems from its non-autoregressive architecture. The accelerated sampling strategy achieves 3$\sim$4$\times$ speed-up beyond DiffSTG ($M$=100, $k$=1). We also find that when $S$ is large, one can increase $k$ for efficiency without loss of performance. See Sectioin~\ref{sec:hyperparameter_study} for more details.

\begin{table}[h]
    \caption{Time cost (by seconds) of diffusion-based models (TimeGrad, CSDI, and DiffSTG) in AIR-GZ ($T_h=12, T_p=12, N = 100$). $S$ is the number of samples.}
    \centering
    \setlength\tabcolsep{10pt}
	\begin{tabular}{l|ccc}
		\toprule
		Method & $S$ = 8 &  $S$ = 16  &   $S$ = 32     \\ 
		\midrule
		TimeGrad \cite{rasul2021autoregressive} &  9.58   &   128.40    & 672.12      \\
        DiffSTG ($M$=100, $k$=1)  &  0.24  &   0.48    & 0.95    \\
        DiffSTG ($M$=40, $k$=1)  &  0.12  &   0.20    & 0.71    \\
        DiffSTG ($M$=40, $k$=2)  &  0.07  &   0.12    & 0.21   \\
        CSDI   &  0.51  &   0.88    & 1.82   \\
		\bottomrule
	\end{tabular}
	\label{tab:time_cost}

\end{table}

\begin{figure*}[!t]
    \centering
    \includegraphics[width=1 \linewidth]{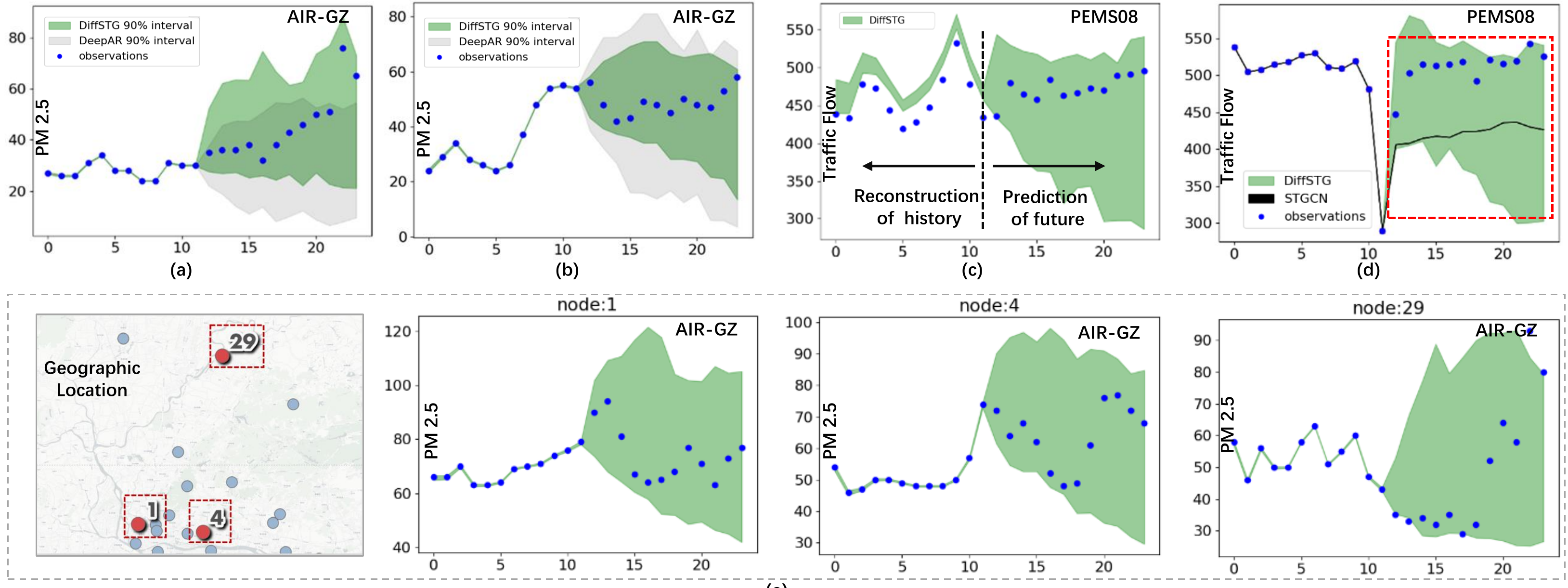} 
    \caption{Example of probabilistic spatio-temporal graph forecasting for air quality and traffic dataset.}
    \label{fig:case_studys}
\end{figure*}

\textbf{Visualization.} We plot the predicted distribution of different methods to investigate their performance intuitively. 
 We choose the AIR-GZ and PEMS08 for demonstration, and more examples on other datasets can be found in Appendix~\ref{appendix:case_sdudy}. We have the following observations: 1) Figure~\ref{fig:case_studys}(a) shows that DiffSTG can capture the data distribution more precisely than DeepAR; 2) in Figure~\ref{fig:case_studys}(b), where the predictions of both DeepAR and DiffSTG cover the observations, DiffSTG provides a more compact prediction interval, indicating the ability to provide more reliable estimations. 3) Note that the model also needs to learn to reconstruct the history in the loss of Eq.~(\ref{eq:masked_condition_loss}), we also illustrate the model's capability in history reconstruction in Figure~\ref{fig:case_studys}(c); 4) Figure~\ref{fig:case_studys}(d) draws the prediction result by a deterministic method STGCN \cite{STGCN-2018} and DiffSTG. In the red box of Figure~\ref{fig:case_studys}(d), the deterministic method fails to give an accurate prediction. In contrast, our DiffSTG model renders a bigger area (which covers the ground truth) in that region, indicating that the data therein is coupled with higher uncertainties. Such ability to accurately provide uncertainty can be of great help for practical decision-making; 5) Moreover, as shown in Figure~\ref{fig:case_studys}(e), we illustrate the estimated distribution of DiffSTG on three stations, to illustrate its spatial dependency learning ability. Compared with station 29, the estimated distribution of station 4 is more similar to station 1, which is reasonable because the air quality of a station has stronger connections with its nearby neighbors. Equipped with the proposed denoising network UGnet, the model is able to capture the ST correlations, leading to more reliable and accurate estimation.

\subsection{Ablation Study}

We conduct an ablation study on the AIR-GZ dataset to verify the effect of each component. We design three variants of DiffSTG and compare them with the full version of DiffSTG. The differences between these four models are described as follows:

\begin{itemize}
    \item  \textbf{(w/o Spatial)}: removing the GNN component in UGnet, i.e., without modeling the spatial correlation.
    \item \textbf{(w/o Temporal)}: removing the TCN component in UGnet, i.e., without modeling the temporal correlation.
    \item \textbf{w/o U-structure}: removing the Unet-based structure in UGnet and only using one TCN for feature extraction.
    \item \textbf{DiffSTG}: the final version of the proposed model.
\end{itemize}

\begin{figure}[!b]
    \centering
    \includegraphics[width=1\linewidth]{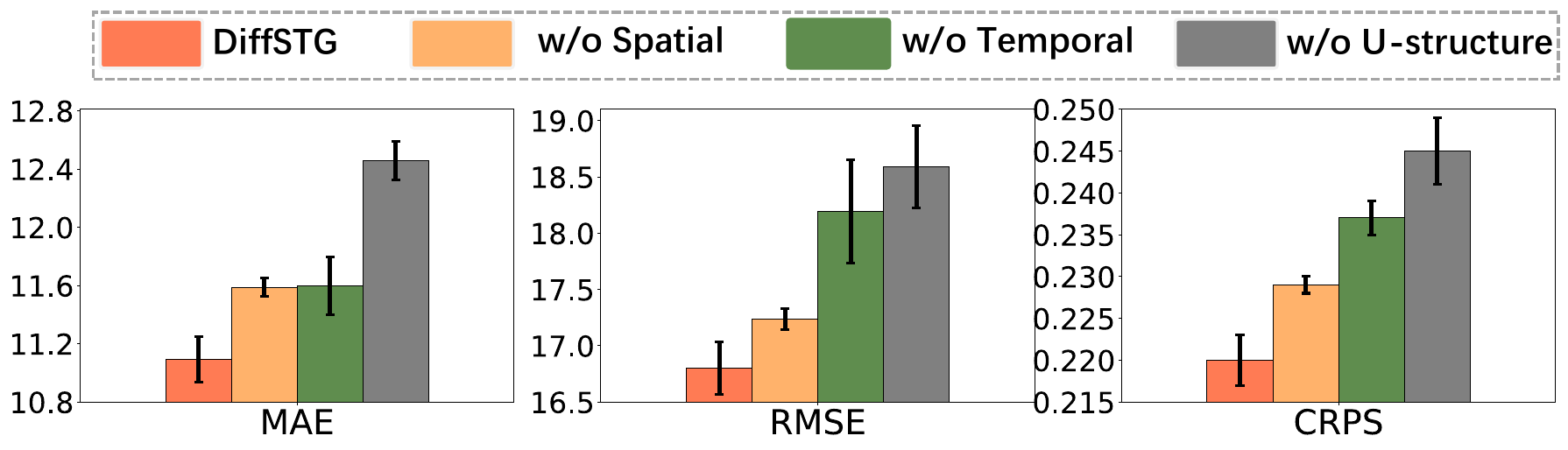} 
    \caption{Ablation study.}
    \label{fig:ablation}
\end{figure}

\par Figure \ref{fig:ablation} illustrates the results. We have the following observations: Firstly, removing the spatial dependency learning in UGnet (w/o Spatial) brings considerable degeneration, which validates the importance of modeling spatial correlations between nodes. Secondly, when turning off the temporal dependency learning in UGnet (w/o Temporal), the performance drops significantly on all evaluation metrics. Thirdly, when detaching the Unet-based structure in UGnet, the performance degrades dramatically, which demonstrates the merits of an Unet-based structure in capturing ST-dependencies at different granularities.

\subsection{Hyperparameter Study} \label{sec:hyperparameter_study}
\par In this section, we examine the impact of several crucial hyperparameters on DiffSTG. Specifically, we report i) the performance on AIR-GZ under different variance schedules (i.e., the combination of $\beta_N$ and diffusion step $N$) and hidden size $C$. ii) influence of kernel size in TCN; iii)  the number of generated samples $S$ and the number of utilized samples $k$ in the proposed  sampling strategy.

\par
\begin{figure}[htbp]
    \centering
    \includegraphics[width=\linewidth]{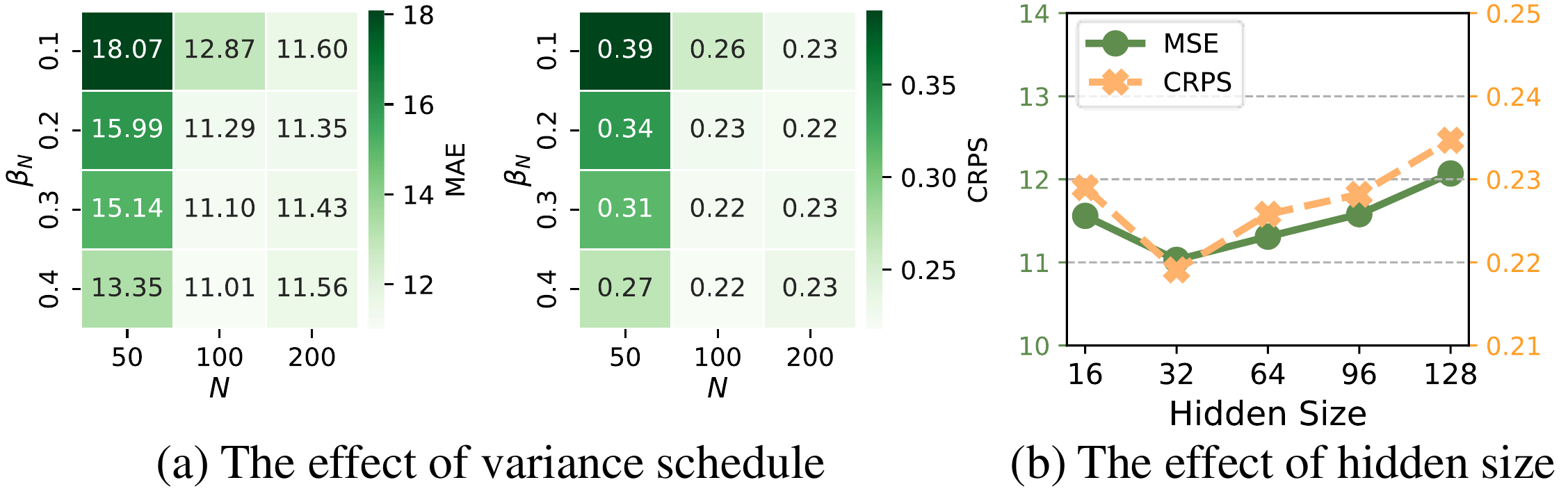} 
    \vspace{-2em}
    \caption{Influence of hyperparameters.}
    \label{fig:hp_sen}
\end{figure}

 \textbf{Effect of the Variance Schedule and Hidden Size.} For different variance schedules $\{\beta_1, \dots, \beta_N\}$, we set $\beta_1 = 0.0001$ and let $\beta_N$ and $N$ be from two search spaces, where $N \in [50, 100, 200]$ and $\beta_N \in [0.1, 0.2, 0.3, 0.4]$. A variance schedule can be specified by a combination of $\beta_N$ and $N$. The results are shown in Figure~\ref{fig:hp_sen}. We note that the performance deteriorates rapidly when $N=50$ and $\beta_N=0.1$. In this case, the result of the forward process is far away from a Gaussian distribution. Consequently, the reverse process starting with Gaussian distribution becomes an inaccurate approximation, which heavily injures the model performance. When $N$ gets larger, there is a higher chance of getting a promising result.

\par Figure~\ref{fig:hp_sen} shows the results of DiffSTG with $N=100$, and $\beta_N=0.4$ vs. different hidden size $C$, from which we observe that the performance first slightly increases and then drops with the increase in hidden size. Compared with the variance schedule, the model's performance is much less sensitive to the hidden size.

\par \textbf{Influence of Kernel Size in TCN.} We report the influence of kernel size ($K =2,3,4,5$) in TCN as in Table~\ref{tab:tcn_influence}. Overall, we can see that the kernel size in TCN has a small impact on the performance of the model. In particular, a kernel size of 3 appears to be a good choice when the input and prediction horizon is 12.

\begin{table}[h]
    \caption{Influence of the kernel size $K$ in TCN on AIR-GZ ($T_h=12, T_p=12, N = 100, \beta_N = 0.1$).}\vspace{-.5em}
    \centering
    \scriptsize
    \setlength\tabcolsep{4pt}
    \resizebox{0.75 \linewidth}{!}{
	\begin{tabular}{ccccc}
		\toprule
		Metric & $K$=2 &  $K$=3  &  $K$=4 & $K$= 5    \\ 
		\midrule
		MAE &  13.85   &   13.38   & 14.02 & 14.86      \\
        RMAE &  19.73   &   19.25    & 19.95 & 20.99      \\
        CRPS &  0.27   &   0.26    & 0.28 & 0.29      \\
		\bottomrule
	\end{tabular}
        }
	\label{tab:tcn_influence}
  \vspace{-.5em}
\end{table}

\begin{figure}[!t]
    \centering
    \includegraphics[width=1 \linewidth]{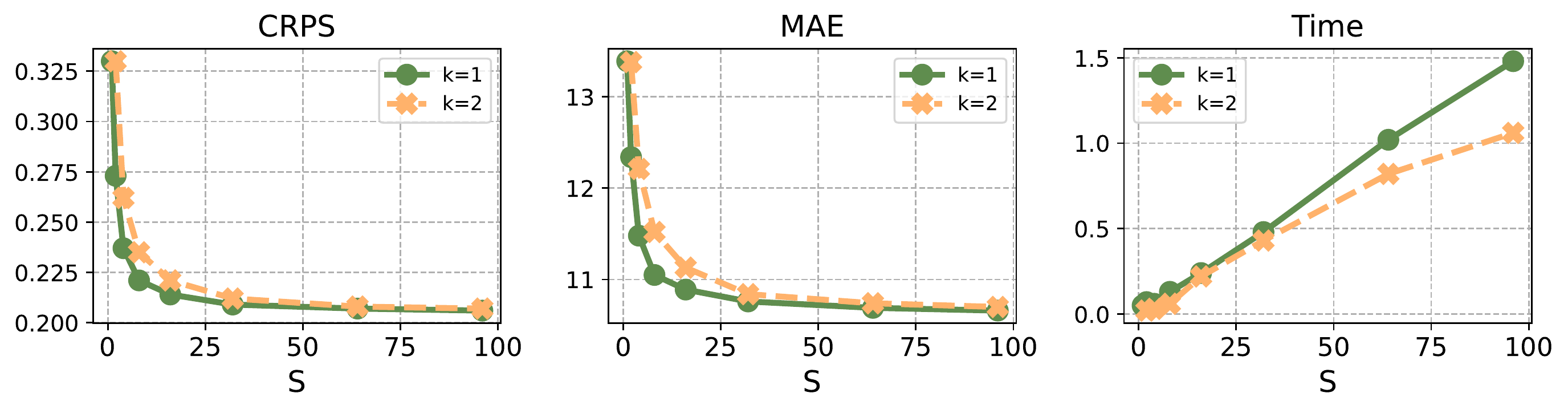} 
    \vspace{-1em}
    \caption{The effect of the number of generated samples and the influence of $k$.}
    \label{fig:n_sample_and_k}
\end{figure}

\textbf{Effect of the Number of Generated Samples $S$.} We investigate the relationship between the number of samples $S$ and the performance in Figure~\ref{fig:n_sample_and_k}. It shows the effect of probabilistic forecasting, as well as the effect on deterministic forecasting. From the figure, we can see that five or ten samples are enough to estimate good distributions. While increasing the number of samples further improves the performance, the improvement becomes marginal over 32 samples.

\textbf{Effect of $k$.} Recall that $k$ is the number of utilized samples in the last few diffusion steps when sampling $S$ samples. We provide the results of $k=1$ and $k=2$ in Figure~\ref{fig:n_sample_and_k}, in which we have several interesting observations: i) When $S$ is large enough (i.e., $S>32$), the performance of $k=2$ is almost the same as $k=1$, and the sample speed of $k=2$ is 1.5 times faster than $k=1$; ii) When the number of reverse diffusion processes (i.e., $S/k$) is settled, a large $k$ can increase sample diversity thus leading to better performance, especially when $S$ is small.

\par In light of the above results, we give the following recommendations for the combination of $S$ and $k$: 1) when $S$ is small, a small $k$ is recommended to increase the sample diversity for better performance; 2) when $S$ is large, one can increase $k$ for efficiency without much lose of the performance.

\subsection{Limitations}
\par Though promising in probabilistic prediction, DiffSTG still has a performance gap compared with current state-of-the-art STGNNs in terms of deterministic forecasting. Table \ref{tab:results_determinstic} shows the deterministic prediction performance of DiffSTG (by averaging 8 in generate samples) and four deterministic methods, including DCRNN \cite{li2018diffusion}, STGCN \cite{STGCN-2018},  STGNCDE \cite{choi2022graph}, and GMSDR \cite{liu2022msdr}. While DiffSTG is competitive with most probabilistic methods, it is still inferior to the state-of-the-art deterministic methods. Different from deterministic methods, the optimization goal of DiffSTG is derived from a variational inference perspective (see details in Appendix.~\ref{appendix:DDPM_detail}), where the learned posterior distribution might be inaccurate when the data samples are insufficient. We have similar observations in other DDPM-based models, such as TimeGrad and CSDI, as shown Table~\ref{tab:results}. 
\par Though, we would like to emphasize a key advantage of probabilistic such as DiffSTG, is its ability to provide probabilistic forecasting. This is the starting point of our focus and an important aspect that even the SOTA GNN-based methods may not be able to offer. We leave improving DiffSTG to surpass those deterministic methods in future work.

\begin{table}[htbp]

\centering
\small
\caption{Comparison with deterministic methods. Lower MAE and RMSE indicate better performance.}
\setlength\tabcolsep{6 pt}

\begin{tabular}{c|cc|cc|cc}
\hline 
\multirow{2}{*}{Method}
& \multicolumn{2}{|c|}{ AIR-BJ}& \multicolumn{2}{|c|}{ AIR-GZ }& \multicolumn{2}{|c}{ PEMS08 }\\
\cline{2-7}
 & MAE & RMSE & MAE & RMSE & MAE & RMSE\\
\hline \hline
DCRNN & \underline{16.99} & \textbf{28.00} & \underline{10.23} & \underline{15.21} & 18.56 & 28.73 \\ \hline
STGCN & 19.54 & 30.51 & 11.05 & 16.54 & 20.15 & 30.14 \\ \hline
STGNCDE & 19.17 & 29.56 & 10.51 & 16.11 & \textbf{15.83} & \underline{25.05} \\ \hline
GMSDR & \textbf{16.60} & \underline{28.50} & \textbf{9.72} & \textbf{14.55} & \underline{16.01} & \textbf{24.84} \\ \hline
DiffSTG & 17.88 & 29.60 & 11.04 & 16.75 & 17.68 & 27.13 \\ \hline
\end{tabular}
\label{tab:results_determinstic}
\end{table}

\section{Related Work}\label{sec:related}

\par  \textbf{Spatio-temporal Graph Forcasting.} Recently, a large body of research has been studied on spatio-temporal forecasting in different scenarios, such as traffic forecasting \cite{STGCN-2018, GraphWaveNet-2019,  guo2021learning, peng2020spatial, ji2022stden} and air quality forecasting \cite{liang2022airformer}. STGNNs have become dominant models in this field, which combine GNN and temporal components (e.g., TCN and RNN) to capture the spatial correlations and temporal features, respectively. However, most existing works focus on point estimation while ignoring quantifying the uncertainty of predictions. To fill this gap, this paper develops a conditional diffusion-based method that couples the spatio-temporal learning capabilities of STGNNs with the uncertainty measurements of diffusion models.

\par \textbf{Score-based Generative Models.} The diffusion model that we adopt belongs to score-based generative models (please refer to Section~\ref{sec:ddpm} for more details), which learn the gradient of the log-density with respect to the inputs, called Stein Score function \cite{hyvarinen2005estimation, vincent2011connection}. At inference time, they use the gradient estimate to sample the data via Langevin dynamics \cite{song2019generative}. By perturbing the data through different noise levels, these models can capture both coarse and fine-grained features in the original data. Which, leads to their impressive performance in many domains, such as image  \cite{ho2020denoising}, audio \cite{kong2020diffwave, chen2020wavegrad}, graph \cite{niu2020permutation} and time series \cite{rasul2021autoregressive, tashiro2021csdi}. 

\par \textbf{Time Series Forecasting.}  Methods in time series forecasting can be classified into two streams. The first one is deterministic methods, including transformer-based approaches \cite{wen2022tstransformers, zhou2021informer, zhou2022fedformer} and RNN-based models \cite{che2018recurrent}. The second one is probabilistic methods \cite{pal2021rnn, salinas2020deepar}, which aims to provide stochastic predictions for future time series. A rising trend in this stream is leveraging the diffusion-based models to improve the performance of probabilistic predictions \cite{rasul2021autoregressive, wang2023diffload, li2023generative, chang2023tdstf}. Though those methods have achieved promising performance, they still lack the ability to model the spatial correlations of different nodes if applied to STG forecasting. The above limitation also  motivates us to develop a more powerful diffusion-based model for STG that can effectively capture spatial correlations. 

\section{Conclusion and Future Work}\label{sec:conclusion}

\par In this paper, we first propose a novel probabilistic framework called DiffSTG for spatio-temporal graph forecasting. To the best of our knowledge, this is the first work that generalizes the DDPM to spatio-temporal graphs. DiffSTG combines the spatio-temporal learning capabilities of STGNNs with the uncertainty measurements of diffusion models. Unlike previous diffusion-based models designed for the image or sequential data, we devise the first denoising network UGnet for capturing the spatial and temporal correlations in STG data. By leveraging an Unet-based architecture to capture multi-scale temporal dependencies and a Graph Neural Network (GNN) to model spatial correlations, UGnet offers a powerful new member to the DDPM denoising network family, tailored for STG data.  Extensive experiments demonstrate the effectiveness and efficiency of our proposed method. 
\par Furthermore, we accelerate the training and inference speed by elaborately designed  non-autoregressive prediction architecture and a sample acceleration strategy. This improvement in efficiency greatly benefits the potential of diffusion-based methods for STG forecasting tasks, making them more viable and appealing for practical applications.

\par Some interesting directions are worth exploring in the future. Since we only use the vanilla GCN in the denoising network for the model's simplicity, it remains open to incorporating more powerful graph neural networks to better capture the ST dependencies in the data. Another direction is to apply DiffSTG to other spatio-temporal learning tasks, such as spatio-temporal graph imputation.


\bibliography{main.bib}
\bibliographystyle{ACM-Reference-Format}


\newpage
\appendix

\section{Appendix}
\subsection{Details of DDPM} \label{appendix:DDPM_detail}
\par We introduce the details of denoising diffusion probabilistic models in this section.

\par Diffusion probabilistic models are latent variable models that consist of two processes, namely the forward process and the reverse process. The forward process is a fixed Gaussian transition process as defined in Eq.~(\ref{eq:forward_process}) and Eq.~(\ref{eq:forward_Gaussian}). The reverse process is a learnable Gaussian transition process defined in Eq.~(\ref{eq:reverse_process}) and Eq.~(\ref{eq:p_sample}). Then, the parameters $\theta$ are learned by minimizing the negative log-likelihood via the variational lower bound (ELBO):
\begin{equation}
    \begin{array}{l}
        \min _\theta \mathbb{E}_{q (\x_0)}\left[ -\log p_\theta (\x_0) \right] \vspace{5pt}  \\
        \leq -\log p_\theta\left(\x_0\right)+D_{\mathrm{KL}}\left(q\left(\x_{1: N} \mid \x_0\right) \| p_\theta\left(\x_{1: N} \mid \x_0\right)\right) \vspace{5pt} \\
        = -\log p_\theta\left(\x_0\right)+\mathbb{E}_{\x_{1: N} \sim q\left(\x_{1: N} \mid \x_0\right)}\left[\log \frac{q\left(\x_{1: N} \mid \x_0\right)}{p_\theta\left(\x_{0: N}\right) / p_\theta\left(\x_0\right)}\right] \vspace{5pt} \\
        = -\log p_\theta\left(\x_0\right)+\mathbb{E}_q\left[\log \frac{q\left(\x_{1: N} \mid \x_0\right)}{p_\theta\left(\x_{0: N}\right)}+\log p_\theta\left(\x_0\right)\right] \vspace{5pt} \\
        =\mathbb{E}_{q(\x_{0:N})}\left[\log \frac{q\left(\x_{1: N} \mid \x_0\right)}{p_\theta\left(\x_{0: N}\right)}\right] := {\mathcal L}_{\rm ELBO}.
        
    \end{array}
\end{equation}
We can further decompose ${\mathcal L}_{\rm ELBO}$ into different terms according to the property of Markov chains:
\begin{equation}
    \small
    \begin{array}{l}
       {\mathcal L}_{\rm ELBO} \vspace{5pt}    \\
       
       = \mathbb{E}_{q(\x_{0:N})}\left[\log \frac{q\left(\x_{1: N} \mid \x_0\right)}{p_\theta\left(\x_{0: N}\right)}\right] \vspace{5pt} \\ 
       
       =\mathbb{E}_q\left[\log \frac{\prod_{n=1}^N q\left(\x_n \mid \x_{n-1}\right)}{p_\theta\left(\x_N\right) \prod_{n=1}^N p_\theta\left(\x_{n-1} \mid \x_n\right)}\right] \vspace{5pt} \\

       =\mathbb{E}_q\left[-\log p_\theta\left(\x_N\right)+\sum_{t=2}^N \log \frac{q\left(\x_n \mid \x_{n-1}\right)}{p_\theta\left(\x_{n-1} \mid \x_n\right)}+\log \frac{q\left(\x_1 \mid \x_0\right)}{p_\theta\left(\x_0 \mid \x_1\right)}\right] \vspace{5pt} \\

       =\mathbb{E}_q\left[\log \frac{q\left(\x_N \mid \x_0\right)}{p_\theta\left(\x_N\right)}+\sum_{t=2}^N \log \frac{q\left(\x_{n-1} \mid \x_n, \x_0\right)}{p_\theta\left(\x_{n-1} \mid \x_n\right)}-\log p_\theta\left(\x_0 | \x_1\right)\right] \vspace{5pt} \\

       =\mathbb{E}_q [\underbrace{D_{\mathrm{KL}}\left(q\left(\x_N \mid \x_0\right) \| p_\theta\left(\x_N\right)\right)}_{\loss_N} \\
       
       + \sum_{t=2}^N \underbrace{D_{\mathrm{KL}}\left(q\left(\x_{n-1} \mid \x_n, \x_0\right) \| p_\theta\left(\x_{n-1} \mid \x_n\right)\right)}_{\loss_{n-1}} \\

       \underbrace{-\log p_\theta\left(\x_0 \mid \x_1\right)}_{\loss_0}].

    \end{array}
\end{equation}
By the property in Eq.~(\ref{eq:forward_property}),  \cite{ho2020denoising} show that the forward process posterior when conditioned on $\x_0$, i.e., $q(\x_{n-1} | \x_{n}, \x_0)$ is tractable, formulated as
\begin{equation}
    q\left(\x_{n-1} \mid \x_n, \x_0\right)=\mathcal{N}\left(\x_{n-1} ; \tilde{\boldsymbol{\mu}}\left(\x_n, \x_0\right), \tilde{\beta}_n \mathbf{I}\right)
\end{equation}
where 
\begin{equation}
    \small
    \tilde{\mu}_n\left(\x_n, \x_0\right)=\frac{\sqrt{{\alpha}_{n-1}} \beta_n}{1-{\alpha}_n} \x_0+\frac{\sqrt{\alpha_n}\left(1-{\alpha}_{n-1}\right)}{1-{\alpha}_n} \x_n,
\end{equation}
and 
\begin{equation}
    \tilde{\beta}_n=\frac{1-{\alpha}_{n-1}}{1-{\alpha}_n} \beta_n.
\end{equation}

\par So far, we can see that each term in $\loss_{\rm ELBO}$ (except for $\loss_0$) calculates the KL Divergence between two Gaussian distributions, therefore they can be computed in closed form. $\loss_N$ is constant that can be ignored in training because $q$ has no learnable parameters and $\x_N$ is a Gaussian noise. \cite{ho2020denoising} models $\loss_0$ using a separate discrete decoder. Especially, the loss term of $\loss_{t}$ ($t \in \{2, \cdots, T\}$), have the following closed form:
\begin{equation}
    \footnotesize
    \mathbb{E}_{\x_0, \epsilon}\left[\frac{\beta_n^2}{2 \Sigma_\theta \alpha_n\left(1-{\alpha}_n\right)}\left\|\epsilon-\epsilon_\theta\left(\sqrt{{\alpha}_n} \x_0+\sqrt{1-{\alpha}_n} \epsilon, n\right)\right\|^2\right],
    \nonumber
\end{equation}
which can be further simplified by removing the coefficient in the loss term, formulated as
\begin{equation}
    \footnotesize
    \mathbb{E}_{\x_0, \epsilon}\left[\left\|\epsilon-\epsilon_\theta\left(\sqrt{{\alpha}_n} \x_0+\sqrt{1-{\alpha}_n} \epsilon, n\right)\right\|^2\right].
    \nonumber
\end{equation}

\subsection{Details of UGnet} \label{appendix:UGnet}

\par We propose a novel denoising network to effectively capture spatio-temporal correlations in STG data, named UGnet. It adopts an Unet-like architecture in the temporal dimension and can also process the graph as the condition. The Unet structure can capture features at different levels because its Convolutional Neural Networks (CNN) kernels gradually merge low-level features into high-level features. Similarly, in the context of spatio-temporal forecasting, we naturally have different granularities in the temporal dimension (e.g., 15 minutes, 30 minutes, and 1 hour). Therefore, an intuitive way is to adopt the idea in Unet that gradually reduce the shape in the temporal dimension and reverse it back so that temporal features at different levels can be well captured. Doing so also brings the model the ability to scale up to large STG.

\par
\begin{figure}[h]
    \centering
    \includegraphics[width=1 \linewidth]{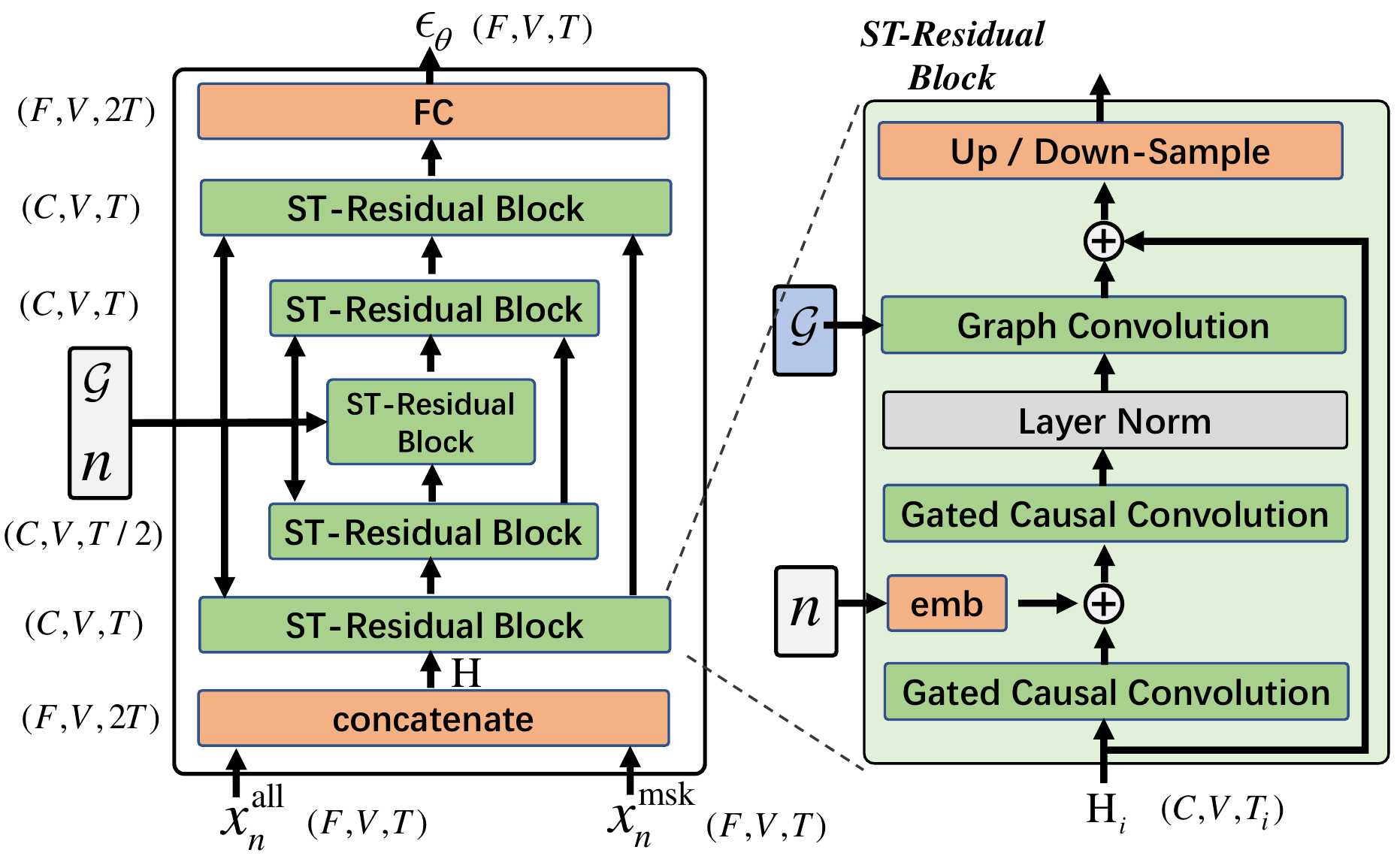} 
    \vspace{-2em}
    \caption{The architecture of denoising network UGnet. It adopts an Unet-like structure to model both spatial and temporal dependencies at different temporal granularities, conditioned on the noise level and given graph structure.}
    \vspace{-1.5em}
    
    \label{fig:epsilon_theta}
\end{figure}

  Specifically, as shown in Figure ~\ref{fig:epsilon_theta}, UGnet $\be_\theta({\mathcal X}^{\rm all} \times {\mathbb R} | {\mathcal X}_{\rm msk}^{\rm all}, \G) \rightarrow {\mathcal X}^{\rm all}$ takes $\xall_{\rm msk}$, $\xall_{n}$, $n$, $\G$ as input, and outputs the denoised noise $\be$.  We first concatenate $\xall_{n} \in \mathbb{R}^{F \times V \times T}$ and $\xall_{\rm msk} \in \mathbb{R}^{F \times V \times T}$ in the temporal dimension to form a new matrix $\xnewall_n \in \mathbb{R}^{F \times V \times 2T}$. UGnet contains several Spatio-temporal Residual Blocks (ST-Residual Block for short) with two types: the down-residual block and the up-residual block. The down-residual blocks gradually reduce the shape in the temporal dimension (i.e., increase the temporal granularity). While the up-residual blocks gradually convert the temporal granularity back to the level that is the same as the input. Both blocks contain the same residual block  that can capture both spatial and temporal correlations in the data with the help of the graph structure.

\par Here, we introduce the details of the ST-Residual block. We first project input $\xall_{n} \in \mathbb{R}^{F \times V \times T}$  into a high-dimensional representation ${\mathbf H} \in {\mathbb R}^{C \times V \times 2T}$ by a linear layer, where $C$ is the projected dimension. Let ${\mathbf H}_{i} \in {\mathbb R}^{C \times V \times T_{i}}$ denote the input of the $i$-th ST-Residual Block, where $T_{i}$ is the length of the time dimension, and $\bmH_0 = \bmH$.

\par \textbf{Temporal Dependence Modeling.} As shown in Figure ~\ref{fig:epsilon_theta}, $\mathbf H_{i}$ is first fed into a Temporal Convolution Network (TCN) \cite{bai2018tcn} for modeling temporal dependence, which is a 1-D gated causal convolution with $K$ kernel size with padding to get the same shape with input.  The convolution kernel $\Gamma_{\mathcal T} \in \mathbb{R} ^ {K \times C_{\rm in}^{\rm t} \times C_{\rm out}^{\rm t}}$ maps the input $\mathbf{H}_{i}$ to two outputs $\mathbf P_{i}$, $\mathbf Q_{i}$ with the same shape ${\mathbf P}_{i}/{\mathbf Q}_{i} \in \mathbb{R}^{C_{\rm out}^{\rm t} \times V \times T_{i}}$. As a result, the temporal gated convolution can be defined as,
\begin{equation}
    \Gamma_{\mathcal T}  ({\mathbf H}_{i})  =  {\mathbf P}_{i} \odot \sigma({\mathbf Q}_{i}) \in \mathbb{R} ^ {C_{\rm out}^{\rm t} \times V \times T_{i}}:= \overline{\bmH}_i,
    \label{eq:tcn}
\end{equation}
where $\odot$ is the element-wise Hadamard product, and $\sigma$ is the sigmoid activation function of GLU. The item $\sigma({\mathbf Q_{i}}) $ can be considered a gate that filters the useful information of $\mathbf P_{i}$ into the next layer. Furthermore, residual connections are implemented among stacked temporal convolutional layers to further exploit the full input times horizon.

\par \textbf{Spatial Dependence Modeling.} Graph convolution network (GCN) is employed to directly extract highly meaningful features and patterns in the space domain. The input of GCN is the node feature matrix, which is reshaped output of the TCN layer in our case, denoted as $ \overline{\bmH}_i \in \mathbb{R}^{V \times C_{\rm in}^{\rm g}} $, where $C_{\rm in}^{\rm g} = T_{i} \times C_{\rm out}^{\rm t})$. A general formulation \cite{survey-GNN-2020} of a graph convolution can be denoted as
\begin{equation}
    \Gamma_{\mathcal G}( \overline{\bmH}_i)  =\sigma \left(\Phi \left(\mathbf{A}_{\rm gcn}, \overline{\bmH}_i\right) \mathbf{W}_i\right),
    \label{eq:gconv}
\end{equation}
where $\mathbf{W}_i \in \mathbb{R}^{C_{\rm in}^{\rm g} \times C_{\rm in}^{\rm g}}$ denotes a trainable parameter and $\sigma$ is an activation function.
$\Phi(\cdot)$ is an aggregation function that decides the rule of how neighbors' features are aggregated into the target node.  In our work,  we do not focus on how to develop an elaborately designed function $\Phi(\cdot)$. Therefore, we use the form in the most popular vanilla GCN~\cite{GCN-kipf2017} that defines a symmetric normalized summation function as $\Phi_{\rm gcn}\left(\mathbf{A}_{\rm gcn}, \overline{\bmH}_i\right) = \mathbf{A}_{\rm gcn} \overline{\bmH}_i,$ where $\mathbf{A}_{\rm gcn}=\mathbf{D}^{-\frac{1}{2}}(\mathbf{A}+\mathbf{I})\mathbf{D}^{-\frac{1}{2}} \in \mathbb{R}^{V\times V}$ is a normalized adjacent matrix. $\mathbf{I}$ is the identity matrix and $\mathbf{D}$ is the diagonal degree matrix with $\mathbf{D}_{ii}=\sum_{j}(\mathbf{A}+\mathbf{I})_{ij}$. 

\par We report the parameter number of probabilistic models below:
\begin{table}[h]
    \vspace{-1em}
    \caption{Parameter number.}
    \vspace{-1em}
    \centering
    \scriptsize
    \setlength\tabcolsep{5pt}
    \resizebox{1 \linewidth}{!}{
	\begin{tabular}{ccccccc}
		\toprule
		Method & TimeGrad &  CSDI  &  DiffSTG &  DeepAR  & LatentODE & MC Dropout\\ 
		\midrule
		  number &  52,423   &   48,289    & 149,305  &  25,474 & 24,073 & 371,393\\
           
		\bottomrule
	\end{tabular}
        }
	\label{tab:parameter_number}
 \vspace{-2.5em}
\end{table}

\subsection{Additional Prediction Examples} \label{appendix:case_sdudy}
\par This section illustrates various probabilistic forecasting examples to show the characteristic of different methods. Note that the scales of the y-axis depend on the stations.

\par
\begin{figure}[htbp]
    \centering
    \includegraphics[width=0.65 \linewidth]{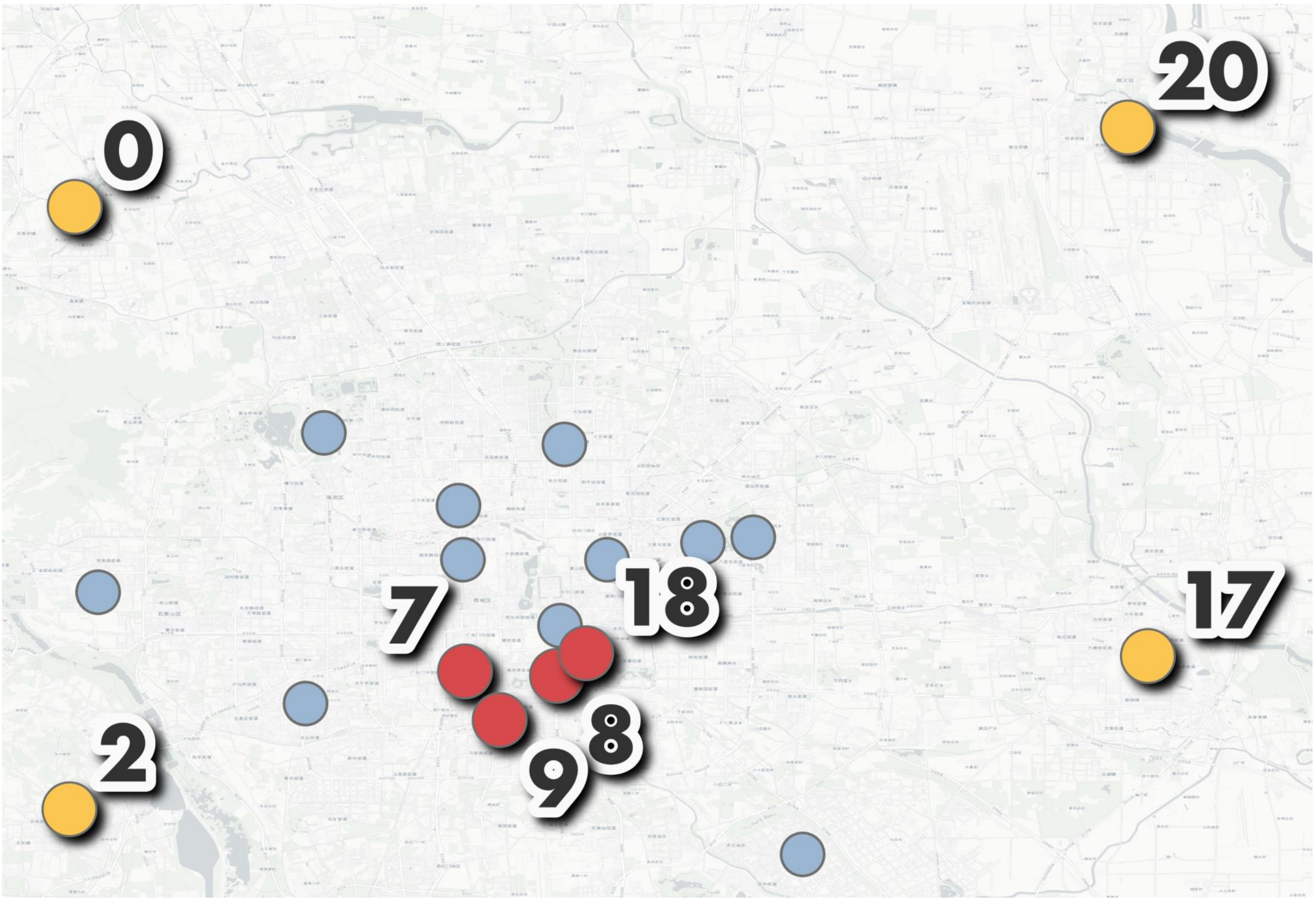} 
    \caption{Geographic distribution of stations on AIR-BJ.}
    \label{fig:AIR_BJ_stations}
    \vspace{-1em}
    
\end{figure}

\begin{figure}[htbp]
    \centering
    \includegraphics[width=1 \linewidth]{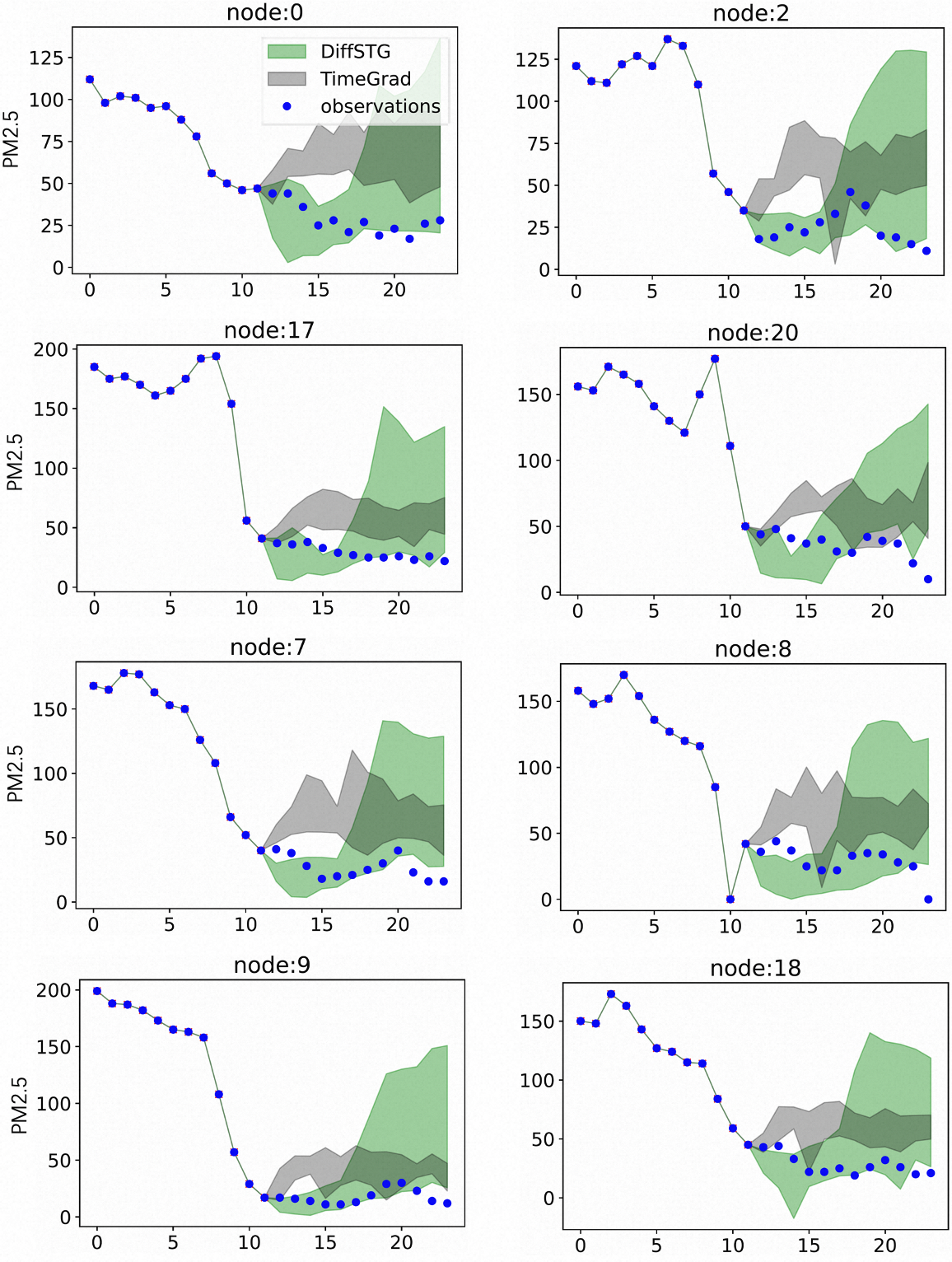} 
    
    \caption{Comparison of probabilistic STG forecasting between TimeGrad and DiffSTG for air quality dataset (AIR-BJ).}
    \label{fig:case_DiffSTG_TimeGrad_AIR_BJ}
\end{figure}

 We compare DiffSTG with  TimeGrad for selected stations of AIR-BJ in Figure~\ref{fig:case_DiffSTG_TimeGrad_AIR_BJ}. The geographic distribution of stations is shown in Figure~\ref{fig:AIR_BJ_stations}. We select two groups of nodes according to their spatial location. Nodes in the first group are far away from each other, including nodes 0, 2, 17, and 20. While nodes in the second group are close to each other, including nodes 7, 8, 9, and 18. 

\par For the comparison in Figure~\ref{fig:case_DiffSTG_TimeGrad_AIR_BJ}, while TimeGrad fails to capture the data distribution, DiffSTG computes reasonable probabilistic forecasting for a majority of the stations.  And we can see that DiffSTG tends to provide similar estimated distribution for stations nearby, which is reasonable since the air quality of a station is strongly correlated with its neighbors. The above examples further illustrate that  DiffSTG can effectively learn the spatial and temporal dependency in STG, thus providing more reliable and accurate estimations than others.

\end{document}